\documentclass[acmtog]{acmart}

\AtBeginDocument{%
  }
    
\setcopyright{acmlicensed}
\acmJournal{TOG}
\acmYear{2024} \acmVolume{43} \acmNumber{4} \acmArticle{79} \acmMonth{7}\acmDOI{10.1145/3658233}

\usepackage{float}
\usepackage{wrapfig}
\usepackage{enumitem}
\usepackage{bm}
\usepackage{multirow}
\usepackage{makecell}
\usepackage{pifont}

\begin{document}

\title{X-SLAM: Scalable Dense SLAM for Task-aware Optimization using CSFD}

\author{Zhexi Peng}
\email{zhexipeng@zju.edu.cn}
\orcid{0000-0003-4342-5263}
\affiliation{
\institution{State Key Lab of CAD\&CG, Zhejiang University}
\city{Hangzhou}
\country{China}
}

\author{Yin Yang}
\email{yin.yang@utah.edu}
\orcid{0000-0001-7645-5931}
\affiliation{
\institution{University of Utah}
\city{Salt Lake City}
\country{United States of America}
}

\author{Tianjia Shao}
\authornote{Corresponding authors: Tianjia Shao (tjshao@zju.edu.cn) and Kun Zhou (kunzhou@acm.org)}
\email{tjshao@zju.edu.cn}
\orcid{0000-0001-5485-3752}
\affiliation{
\institution{State Key Lab of CAD\&CG, Zhejiang University}
\city{Hangzhou}
\country{China}
}

\author{Chenfanfu Jiang}
\email{chenfanfu.jiang@gmail.com}
\orcid{0000-0003-3506-0583}
\affiliation{
\institution{University of California, Los Angeles}
\city{Los Angeles}
\country{United States of America}
}

\author{Kun Zhou}
\authornotemark[1]
\email{kunzhou@acm.org}
\orcid{0000-0003-4243-6112}
\affiliation{
\institution{State Key Lab of CAD\&CG, Zhejiang University}
\city{Hangzhou}
\country{China}
}

\renewcommand{\shortauthors}{Peng et al.}

\begin{abstract}
We present X-SLAM, a real-time dense differentiable SLAM system that leverages the complex-step finite difference (CSFD) method for efficient calculation of numerical derivatives, bypassing the need for a large-scale computational graph. The key to our approach is treating the SLAM process as a differentiable function, enabling the calculation of the derivatives of important SLAM parameters through Taylor series expansion within the complex domain. Our system allows for the real-time calculation of not just the gradient, but also higher-order differentiation. This facilitates the use of high-order optimizers to achieve better accuracy and faster convergence. Building on X-SLAM, we implemented end-to-end optimization frameworks for two important tasks: camera relocalization in wide outdoor scenes and active robotic scanning in complex indoor environments. Comprehensive evaluations on public benchmarks and intricate real scenes underscore the improvements in the accuracy of camera relocalization and the efficiency of robotic navigation achieved through our task-aware optimization. The code and data are available at \url{https://gapszju.github.io/X-SLAM}.
\end{abstract}




\begin{teaserfigure}
  \center
  \includegraphics[width = \textwidth]{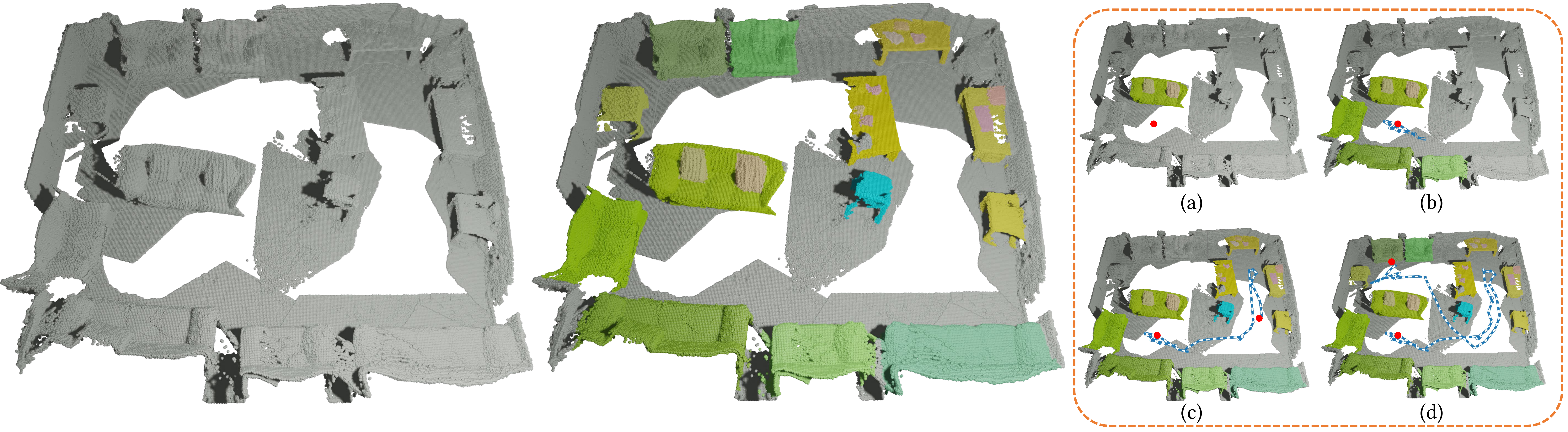}
  \Description[Figure 1: Real-time robot active scanning and reconstruction with semantic segmentation based on our X-SLAM system]{Left: reconstruction result, middle: semantic segmentation result, right: robot scanning process. We utilize CSFD to integrate the differentiable dense SLAM system with a neural network, and allow the robot to accomplish active scanning with semantic guidance.}
  \caption{Real-time robot active scanning and reconstruction with semantic segmentation based on our X-SLAM system. With X-SLAM, robots can carry out automatic navigation and scanning within an unknown environment (left), and obtain a reconstruction with semantic segmentation (middle). The scanning process is presented on the right. We propose the first real-time differentiable dense SLAM system utilizing CSFD. By integrating it with a neural network, we facilitate robot active scanning and scene comprehension with semantic awareness.}
  \label{fig:teaser}
\end{teaserfigure}

\maketitle

\section{Introduction}\label{sec:intro}
Simultaneous Localization and Mapping (SLAM) is a key technology in a variety of fields including augmented reality, robot navigation, autonomous driving, etc. While the primary objective of most SLAM algorithms is to reconstruct 3D scenes through RGB/RGBD cameras~\cite{newcombe2011kinectfusion,whelan2015ElasticFusion,dai2017bundlefusion}, they have been deeply integrated with an array of downstream tasks, where the 3D geometry of the scene is more or less an ``intermediate product''. For instance, \citet{liu2018object} and \citet{gottipati2019deep} combine SLAM with the robot navigation task based on scene understanding. \citet{wu2021object} present a SLAM-based framework for complex robotic manipulation and autonomous perception tasks. \citet{mccormac2017SemanticFusion} and \citet{zhang2018Semantica} also map environments containing semantics for robot intelligence and user interaction. \citet{stenborg2018Longterm} propose a cross-seasonal localization task after the SLAM 3D reconstruction.

Typically, these methods treat the SLAM system and the downstream task as separate modules. That said, the SLAM system first produces the results of 3D reconstruction and camera poses, and the downstream application makes the subsequent decisions based on the SLAM outputs. The quality, reliability, robustness, and performance of those tasks rely on the accuracy of the SLAM result. 
Once errors accumulate during the scanning process, the downstream task has to be processed based on the erroneous SLAM results. More recently, researchers propose to express the SLAM system as a differentiable function~\cite{jatavallabhula2020GradSLAM}, which can be combined with downstream tasks in an end-to-end manner allowing the task-based error signals to be back-propagated all the way to the raw sensor observations. In this way, the 3D reconstruction can be learned to optimize the task performance. For instance, SLAM-net~\cite{karkusDifferentiableSLAMnetLearning2021} replaces the SLAM modules with neural networks to enable differentiability. Its generalization ability is naturally limited to seen scenes. As a seminal work, $\nabla$SLAM~\cite{jatavallabhula2020GradSLAM} first presents a fully differentiable 3D dense SLAM system that thoroughly addresses the non-differentiable steps in traditional SLAM frameworks by offering mathematically differentiable approximations while preserving accuracy. This paradigm relies on the computation graph~\cite{NEURIPS2019_9015}. The computational graph of all tracking/mapping functions needs to be maintained at every frame, leading to a continuous increase in memory consumption. Based on the public code, the SLAM procedure will crash after 60 iterations on PointFusion with image resolution of 640$\times$480. The efficiency is not sufficient for the practical application due to the backward propagation cost in large computation graphs. These limitations restrict the capability of differentiable SLAM on real applications, where long-duration and real-time scanning is common.   

In this paper, we present X-SLAM, a real-time and differentiable dense SLAM system that allows task-oriented optimization of SLAM parameters based on the loss backpropagation. Inspired by \cite{jatavallabhula2020GradSLAM}, we address several fundamental challenges to make differentiable SLAM practical. First, we need to find a memory-efficient way to compute the differentials, so as to avoid the memory problem caused by the increasing computation graph. Second, the differential computation should be in real time to match the frame rate of SLAM. 
Third, in many real-world tasks, the first-order optimization may get stuck in the local minimum, prohibiting high accuracy of the end-to-end optimization. Our method should support the computation of high-order differentials.  

We employ the complex-step finite difference (CSFD) method~\cite{shen2021high} to make a fully differentiable SLAM system for task-aware optimization, avoiding the expensive cost of computation graphs. CSFD calculates the differential by Taylor series expansion in the complex domain. We treat the SLAM procedure as a function $f$ and represent the SLAM input $x$ as a complex number $x^*$ with a small non-zero imaginary perturbation, and the differential can be directly obtained by taking the imaginary part of $f(x^*)$ without the need of maintaining the computation graph. In order to achieve real-time computation of a differential, we utilize CSFD to automatically track the differentials of all variables w.r.t. perturbed inputs during the SLAM process, which naturally eliminates the backpropagation step in the computation graph. CSFD can also be generalized using high-order complex numerics to compute high-order differentials such as Hessian matrix. This allows us to exploit more powerful optimizers to achieve high accuracy.

We demonstrate X-SLAM in two important tasks: robot active scanning and reconstruction of indoor scenes, and camera relocalization in outdoor scenes. For both tasks, we develop task-aware objective functions and optimization pipelines to optimize SLAM parameters based on the loss backpropagation. We conduct careful comparisons on diverse public benchmark datasets and evaluate the performance of our approach in multiple challenging real scenes. The comparisons show that the X-SLAM based framework outperforms the state-of-the-art methods qualitatively and quantitatively.

\section{Related Work}\label{sec:related}
\paragraph{Task-aware SLAM} 
SLAM within a commodity RGBD camera has been extensively studied. Many excellent RGBD SLAM algorithms~\cite{newcombe2011kinectfusion, mur-artal2015ORBSLAM, whelan2015ElasticFusion} have been proposed and the latest SLAM methods~\cite{dai2017bundlefusion, xu2022HRBFFusion} can achieve impressive reconstruction and localization accuracy. In recent years, higher-level information such as semantics in the SLAM process has received increasing attention. \citet{huang2021Supervoxel} propose a novel convolution operation on supervoxels which can fuse the multi-view 2D features and 3D features during the online reconstruction. \citet{liu2022INSConv} achieve online 3D joint semantic and instance segmentation based on an incremental sparse convolutional network. On the other hand, \citet{SLAM++Salas2013} exploit an object database to improve the accuracy of localization and reconstruction quality. \citet{ObjectMatchGümeli2023} present a semantic and object-centric camera pose estimator for RGB-D SLAM pipelines to complete robust registration. \citet{VirtualCorrespondenceMa2022} present a novel concept called virtual correspondences to establish geometric relationships between little overlap images. However, they do not apply those segmentations toward explicit downstream tasks. \citet{jatavallabhula2020GradSLAM} propose a fully differentiable SLAM framework, called $\nabla$SLAM, based on the computational graph and points out that the combination of differentiable SLAM and gradient-based optimization can achieve task-aware SLAM systems. X-SLAM is inspired by $\nabla$SLAM. We avoid the fundamental limitations of $\nabla$SLAM, which is based on automatic differentiation (AD), and propose CSFD-based differentiation to achieve more efficient and memory-economic differentiation.

\paragraph{Derivatives evaluation}
\emph{Forward difference} (FD) is commonly used numerical differentiation method. A drawback of FD is that the numerical errors introduced during the subtraction process can significantly impact the accuracy. CSFD addresses numerical stability issues by extending the calculations to the complex domain and adding perturbations to the imaginary part. Extending the function to the \emph{dual domain} can also achieve similar effects. In the dual domain, a \emph{dual number} $d=x+y\epsilon$ has the real part $x$ and a dual part $y\epsilon$. Dual numerics define $\epsilon^2 = 0$ so $f(x_0+h\epsilon) = f(x_0) + f'(x_0)\cdot h\epsilon$ because the higher-order terms of $\epsilon$ are vanished. High-order differentials can be computed by \emph{hyper-dual number}~\cite{fike2011development, cohen2016application, Aguirre2020MultiZ}. Different from CSFD, the derivative is extracted by setting $h=1$ as: $f'(x_0)=\mathtt{Du}(f(x_0+\epsilon))$, so the dual number method is step-size-independent and exact to machine precision. However, \citet{Aguirre2020MultiZ} show that CSFD can achieve around $10^{-8}$ relative error for step size $h = 10^{-8}$, and it has better compatibility with other computational libraries such as Eigen and cuBLAS. Therefore, we use CSFD to compute differentials.

\paragraph{Camera relocalization} 
Camera relocalization in known scenes is a 3D geometry task closely related to SLAM. Most existing methods accomplish relocalization through the process of feature detection~\cite{sattler2011Fast}, description~\cite{detone2018SuperPoint}, and matching~\cite{sarlin2020SuperGlue}. For example, \citet{sarlin2019Coarse, sarlin2021Back} accomplishes relocalization through feature matching on 2D images from coarse to fine. \citet{brachmann2018DSAC} predicts dense correspondences between the input image and the 3D scene space, using differentiable RANSAC to optimize the camera pose. On the other hand, some research indicates that neural networks can directly predict camera poses. PoseNet~\cite{kendall2016PoseNet} is the first method to directly regress the camera pose from a single image and recent works use attention and transformer to improve accuracy~\cite{wang2019AtLoc, li2022GTCaR}. However, these methods rely on color features, making it difficult to complete relocalization in areas with missing or repetitive textures. Although some methods~\cite{brachmann2021visual, tang2021Learning} use depth images to increase geometric constraints, their accuracy is still limited. Thanks to X-SLAM, we reconstruct from query images and reference images respectively, calculate errors, build a differentiable objective function to optimize the pose parameters and achieve high-accuracy camera relocalization.

\paragraph{Robot active scanning}
Active scanning and online reconstruction for robots in unknown environments is one of the important tasks of SLAM~\cite{callieri2004roboscan, zeng2020view}. \citet{gonzalez2002Navigatio} propose an exploration approach for selecting a new goal according to the maximum map expansion. Building upon this idea, many efforts have been made based on reconstructed maps to develop exploration strategies. For example, RH-NBV~\cite{bircher2016Receding} compute a random tree where branch quality is determined by the amount of unmapped space that can be explored and the first eddge is executed at every planning step. \citet{xu2017Autonomous} harness a time-varying tensor field conforming to the progressively reconstructed scene to guide robot movement. On the other hand, semantic information is also applied to provide high-level guidance. \citet{liu2018object} introduce an approach that interleaves between object analysis to identify the next best object for global exploration, and object-aware information gain analysis to plan the next best view for local scanning. \citet{zheng2019active} propose an online estimated discrete viewing score field (VSF) parameterized over the 3D space of 2D location and azimuth rotation to guide the robot. These methods depend on the mapping quality and tracking accuracy of SLAM. When errors accumulate, obvious mistakes will happen. We develop a novel active scanning approach which can accomplish both activate exploration and semantic segmentation, based on our differentiable SLAM system and object recognition with deep learning.

\section{Complex-step Finite Difference}\label{sec:csfd}
To make the paper more self-contained, we start with a brief review of CSFD. We refer the reader to related literature e.g., see~\cite{martins2003complex} for a more thorough discussion and analysis.

Given a differentiable function $f:\mathbb{R}\rightarrow\mathbb{R}$, we apply a small perturbation $h$ at $x = x_0$. The perturbed function can be Taylor expanded as:
\begin{equation}\label{eq:taylor}
f(x_0+h) = f(x_0)+f'(x_0)\cdot h + \mathbf{O}(h^2),
\end{equation}
which leads to the forward difference of: 
\begin{equation}\label{eq:forward_difference}
f'(x_0)=\frac{f(x_0+h)-f(x_0)}{h} + \mathbf{O}(h) \approx\frac{f(x_0+h)-f(x_0)}{h}.
\end{equation}
The numerator of Eq.~\eqref{eq:forward_difference} evaluates the difference between two quantities of similar magnitude as we often want $h$ to be as small as possible for a better approximation. Subtraction is numerically unstable~\cite{higham2002accuracy}, and the so-called \emph{subtraction cancellation} could occur when $f(x_0+h)$ and $f(x_0)$ become nearly equal to each other. Subtraction between them would eliminate many leading significant digits, and the result after the rounding could largely deviate from the actual value of $f(x_0+h)-f(x_0)$. Because of this limitation, the finite difference is poorly suited for estimating derivatives of functions when high accuracy is needed.

CSFD offers a different approach to numerical derivatives. Instead of perturbing the function in the real domain, CSFD makes the perturbation an imaginary quantity, i.e., $hi$, and the corresponding Taylor expansion becomes:
\begin{equation}\label{eq:complex_taylor}
f^*(x_0+hi)=f^*(x_0)+f^{*'}(x_0)\cdot{hi}+\mathbf{O}(h^2).
\end{equation}
Here $f^*$ suggests the function is ``promoted'' to the complex domain, and its first-order derivative can be approximated by extracting the imaginary parts of Eq.~\eqref{eq:complex_taylor}:
\begin{equation}\label{eq:csfd}
f'(x_0)=\frac{\mathtt{Im}\big(f^*(x_0+hi)\big)}{h}+\mathbf{O}(h^2)\approx\frac{\mathtt{Im}\big(f^*(x_0+hi)\big)}{h}.
\end{equation}
Compared with Eq.~\eqref{eq:forward_difference}, it is noted that CSFD has a second-order convergence ($\mathbf{O}(h^2)$) as $h$ approaches zero, and it does not suffer from the subtraction cancellation. In other words, we can make $h$ as small as needed to fully suppress the approximate error: when $h$ is at the order of $\sqrt{\epsilon}$ ($\epsilon$ is the machine epsilon), CSFD becomes as accurate as the analytic derivative for a given floating point system.

CSFD can be generalized to tackle high-order differentiation by further lifting the perturbation to be a \emph{multicomplex} quantity. A multicomplex number is defined recursively: the base cases are the real number $\mathbb{R}$ and the regular complex number $\mathbb{C}$, which are considered as the zero- and first-order multicomplex sets $\mathbb{C}^0$ and $\mathbb{C}^1$. The complex number set $\mathbb{C}^1$ extends the real set by adding an imaginary unit $i$ as: $\mathbb{C}^1=\{x+yi | x, y\in\mathbb{C}^0\}$, and the multicomplex number up to an order of $n$ is defined as:
\begin{equation}\label{eq:multi_complex}
\mathbb{C}^n=\left\{z_1+z_2 i_n | z_1, z_2\in\mathbb{C}^{n-1}\right\}.
\end{equation}
Following the derivation in~\cite{lantoine2012using}, the Taylor series expansion of $f^{\star}$ under a multicomplex perturbation is:
\begin{equation}\label{eq:mc_taylor}
\begin{split}
&f^{\star}(x_0+hi_1+\cdots+hi_n)=f^{\star}(x_0)+f^{\star(1)}(x_0)\cdot h\sum_{j=1}^ni_j \\
&+\frac{f^{\star(2)}(x_0)}{2}\cdot h^2 \left(\sum_{j=1}^ni_j\right)^2
+\cdots+\frac{f^{\star(n)}}{n!}\cdot h^n \left(\sum_{j=1}^ni_j\right)^n +\cdots.
\end{split}
\end{equation}
Here, $f^{\star(n)}$ is the $n$-th-order derivative of $f^{\star}$. $\left(\sum i_j\right)^k$ can be expanded following the multinomial theorem, which contains products of mixed $k$ imaginary directions for the $k$-th-order term. Because $\left(\sum i_j\right)^k\neq\left(\sum i_j\right)^l$ for $k \neq l$, Eq.~\eqref{eq:mc_taylor} allows us to approximate an arbitrary-order derivative by extracting the corresponding imaginary combination, just as we did in Eq.~\eqref{eq:csfd}. 
For instance, elements of the Hessian matrix (of a function $f(x,y): \mathbb{R}^2\rightarrow\mathbb{R}$) can be easily obtained as:
\begin{equation}\label{eq:csfd_hessian}
\left\{
\begin{array}{l}
  \displaystyle \frac{\partial^2f(x,y)}{\partial x^2} \approx \frac{\mathtt{Im}^{(2)}\big(f(x+hi_1+hi_2, y)\big)}{h^2},\\

  \displaystyle \frac{\partial^2f(x,y)}{\partial y^2} \approx \frac{\mathtt{Im}^{(2)}\big(f(x, y+hi_1+hi_2)\big)}{h^2},\\

  \displaystyle \frac{\partial^2f(x,y)}{\partial x \partial y} = \frac{\partial^2f(x,y)}{\partial y \partial x} \approx \frac{\mathtt{Im}^{(2)}\big(f(x+hi_1, y+hi_2)\big)}{h^2}.
\end{array}\right.
\end{equation}

CSFD lays out the foundation of our X-SLAM framework serving as the primary differentiation modality of the pipeline. Since the complex numerics are only for gradient/differentiation calculation, the imaginary part of the computation is always small (i.e., corresponding to a small variation induced by the input perturbation $hi$). In this situation, it is unnecessary to perform general-purpose complex numerics and operations. For instance, the multiplication and division between two complex numbers $z_1=a_1+b_1i$ and $z_2=a_2+b_2i$ can be simplified as:
\begin{equation}\label{eq:csfd_simplify_multiplication}
     z_1z_2  = (a_1 a_2-b_1 b_2)+(a_1 b_2+a_2 b_1 )i \approx a_1 a_2+(a_1 b_2+a_2 b_1 )i, 
\end{equation}
and
\begin{equation}\label{eq:csfd_simplify_division}
\frac{z_1}{z_2} = \frac{a_1 a_2+b_1 b_2}{a_2^2+b_2^2}+\frac{b_1 a_2-a_1 b_2}{a_2^2+b_2^2}i
         \approx\frac{a_1}{a_2} +\frac{b_1 a_2-a_1 b_2}{a_2^2}i.    
\end{equation}

We use such a simplification strategy for most elementary functions such as exponential functions, logarithmic functions, power functions, trigonometric functions, etc., which are used in the SLAM pipeline by discarding high-order products between two imaginary quantities. We encapsulate these implementations as a complex library using \textsf{C++} on CPU and \textsf{CUDA} on GPU by overloading most floating-point operators to enable accelerated complex number computations.

To better understand computational overhead induced by complex promotion, we conduct a simple experiment to compare CSFD with AD and FD. We randomly sample $x$ within the range of $[0,1]$ and calculate the first-order numerical derivative of $f(x)=(e^x+x^3+x) / (x + 1)$ at $x$. The process is repeated one million times on the CPU. AD is implemented in \textsf{Python} with PyTorch, FD is implemented in \textsf{C++} with STD library and CSFD is implemented in \textsf{C++} with our complex library. We report the total time, memory usage, and relative error in Table \ref{table: comparison_ad_fd_csfd}. The relative error is defined as:
\begin{equation}
    e_{relative}=|\frac{v_0-v}{v}|,
\end{equation}
where $v$ is the analytic value, and $v_0$ is the numerical derivation. We regard the result of AD as the analytic value. The results indicate that both FD and CSFD outperform AD method in terms of time and memory performance because they do not require maintaining the computational graph. Although FD has a faster forward speed, the time difference or the performance difference between CSFD and FD is not significant because FD needs to compute the original function twice, and the accuracy of CSFD far exceeds that of FD.

It should also be noted that the time performance also depends on the input and output dimensions of the function. The derivative of a parameter needs one forward pass of the function, and the total computational cost linearly depends on the number of parameters. To this end, we slightly change the previous example by making the input of $f$ a 10-dimension vector such that $f(y)=(e^y+y^3+y) / (y + 1)$, for $y=||x||$, and $x\in\mathbb{R}^{10}$. The total calculation time for AD, FD, CSFD increases to 137.72s, 1.05s, and 2.64s respectively. In this case, inverse AD can obtain derivatives for all parameters with one backpropagation process, making it a better choice for applications that differentiate with respect to a large number of parameters, such as deep learning.
{\small \fontfamily{ppl}\selectfont
\begin{table}[t]
    \caption{\textbf{Comparison between inverse AD, FD and CSFD.} The timing information of the forward computation is also given in the parentheses. The results show that CSFD achieves better accuracy compared with FD, and outperforms inverse AD in computational efficiency}
    \renewcommand\arraystretch{1.3}
    \begin{tabular}{c | c | c | c }
    \toprule[1pt]
    Method  & Time                & Memory     &  Relative error   \\ \hline
    AD      &  120.2s (39.8s)    & 1.53MB      & -                 \\
    FD      &  163.8ms (81.9ms)  & 14KB        & 6.49e-02          \\
    CSFD    &  227.2ms (227.2ms) & 32KB        & 8.93e-08          \\
    \bottomrule[1pt]
    \end{tabular}
    \label{table: comparison_ad_fd_csfd}
\end{table}
}

\section{X-SLAM}\label{sec:xslam}
In this section, we walk through our X-SLAM pipeline with details. Similar to existing dense SLAM systems, Our X-SLAM consists of surface measurement, prediction, and camera pose estimation. With the help of CSFD, we aim to calculate derivatives at each step with improved scalability and efficiency. 

As long as the imaginary perturbation is applied to the parameter of interest, we can easily obtain the corresponding derivative by extracting the imaginary part of the result (i.e., Eq.~\eqref{eq:csfd}). There are a few non-smooth computations that could potentially lead to ill-defined gradients, and some special treatments will be needed. Unless the gradient is needed, we could discard and ignore the real-part computation as in~\cite{luo2019accelerated}. 

\subsection{X-KinectFusion}\label{subsec:xkinect_fusion}
Kinect Fusion (KF)~\cite{newcombe2011kinectfusion} is one of the most used dense SLAM algorithms. When an RGBD frame $\{ I_k, D_k \}$ is received from the sensor at time $k$, KF estimates the current camera pose $\mathsf{T}_{g, k}$ and updates to the global fusion $\mathcal{M}$. 

\paragraph{Surface measurement}
Given the camera calibration matrix $\mathsf{K}$, the depth map $D_k$ is obtained by applying a bilateral filter~\cite{tomasi1998bilateral} to the raw depth map for noise reduction. Each image pixel $u = [u_x, u_y]^\top$ corresponds to a vertex in the current camera frame:
\begin{equation}\label{eq:vertex}
V_k(u) = D_k(u)\mathsf{K}^{-1}\dot{u}.
\end{equation}
Here $V_k$ is the vertex map in the camera's local coordinate. $\dot{u}$ is the homogeneous vector $\dot{u} := (u^\top|1)^\top$. The normal vector of the vertex can be computed as:
\begin{equation}\label{eq:normal}
    N_k(u) = \frac{[V_k(u^{x+}) - V_k(u)]\times[V_k(u^{y+}) - V_k(u)]}{\left\|[V_k(u^{x+}) - V_k(u)]\times[V_k(u^{y+}) - V_k(u)]\right\|}.
\end{equation}
Here $u^{x+} = [u_x + 1, u_y]^\top$ and $u^{y+} = [u_x, u_y+1]^\top$. This computation is trivially parallelizable on the GPU as we launch one \textsf{CUDA} thread to process each pixel.

In general, the procedure of surface measure is a map of:
\begin{equation}\label{eq:surface_measure}
(V_k, N_k)\leftarrow \mathtt{surface}_{-}\mathtt{measure}(u, \mathsf{K}, D_k).
\end{equation}
We typically regard $\mathsf{K}$ accurate and constant and want to understand how the depth errors are propagated to $V_k(u)$ and $N_k(u)$, as well as in the follow-up computations. This information allows us to control the depth noise according to vertex and normal maps. Therefore, we apply the CSFD promotion to $D_k(u)$ as $D_k^*(u)$, making both $V_k(u)$ and $D_k(u)$ complex, i.e., $V^*_k(u)$ and $N^*_k(u)$ in Eqs.~\eqref{eq:vertex}~and~\eqref{eq:normal}:
\begin{equation}\label{eq:csfd_depth}
    D_k^*(u) = D_k(u) + hi,
\end{equation}
where we set $h = 1e-8$ to match the CSFD approximation error to $\epsilon$. The partial derivative of $V_k$ and $N_k$ w.r.t. depth variation can be obtained as:
\begin{equation}\label{eq:pvnpd}
    \frac{\partial V_k(u)}{\partial D_k(u)} = \frac{\mathtt{Im}\big(V^*_k(u)\big)}{h}, \; \text{and}\; \frac{\partial N_k(u)}{\partial D_k(u)} = \frac{\mathtt{Im}\big(N^*_k(u)\big)}{h}.
\end{equation}
KF computes depth maps of different resolutions for multi-level ICP alignment. A coarser level depth is computed by applying Gaussian smoothing over $D_k(u)$. The perturbed $hi$ is also Gaussian smoothed.

\paragraph{Ray casting}
Ray casting is needed for surface prediction. Given the camera pose $\mathsf{T}_{g, k}$, a ray is cast through the center of each pixel $u$ in the image: 
\begin{equation}
r(\alpha, u) = \alpha \mathsf{R}_{k} \mathsf{K}^{-1} \dot{u} + t_{k}, \; \text{where}\, \mathsf{T}_{g,k} = 
\left[
\begin{array}{cc}
     \mathsf{R}_{k} &  t_{k}\\
     0 & 1 
\end{array}
\right]\in\mathbb{SE}(3).
\end{equation}
We trace along all the rays by increasing the parameter $\alpha$ each time by $\Delta \alpha$: $\alpha_{m+1} \leftarrow \alpha_m + \Delta \alpha$. This corresponds to a point in the global frame $v_m = r (\alpha_m, u)$ whose TSDF value $F(v_m)$ is obtained via trilinear interpolation. A zero crossing is then confirmed once we have $F(v_{m+1}) < 0 $ and $F(v_m) > 0$ as:
\begin{equation}
    \alpha^0 = \alpha_{m} - \frac{\Delta \alpha F(v_{m})}{F(v_{m+1}) - F(v_m)},
\end{equation}
which corresponds to a vertex $V_{g,k}(u)$ on the surface. Then the normal vector at this surface point can be computed as: $N_{g,k}(u) = {\nabla F(r(\alpha^0, u))}/{\|\nabla F(r(\alpha^0, u))\|}$.

This procedure suggests that the prediction of a surface point depends on the camera pose $\mathsf{T}_{g, k}$:
\begin{equation}\label{eq:surface_prediction}
(V_{g,k}, N_{g,k})\leftarrow \mathtt{surface}_{-}\mathtt{prediction}(u, \mathsf{T}_{g, k}).
\end{equation}

To evaluate its differentiation w.r.t. the camera pose, we apply the complex promotion at the exponential map of the camera rotation:
\begin{equation}\label{eq:lie}
\mathsf{T}^*_{g, k} = \exp{(\xi^*_{k})} = \exp{\left(\left[
\begin{array}{c}
\phi^*_k\\
\rho^*_k
\end{array}
\right]
\right)}=
\left[
\begin{array}{cc}
\exp{(\phi^{*}_{k})} & \widetilde{\mathsf{R}}^*_{k}\rho^*_{k}\\
0 & 1
\end{array}
\right].
\end{equation}
We then use CSFD to compute the derivative w.r.t. $\xi_k^*$ after promoting it to $\mathbb{C}^6$. $\phi^{*}_{k} \in \mathbb{C}^3$ is the promoted rotation vector such that:
\begin{equation}\label{eq:rotvector}
\mathsf{R}^*_k = \exp{(\phi^{*}_k)} = \cos \theta^*\mathsf{Id} + (1 - \cos \theta^*) a^* a^{*\mathsf{H}} + \sin \theta^* [a^*].
\end{equation}
$\mathsf{Id}$ is the 3 by 3 identity matrix. $\theta^*$ is the complex-perturbed rotation angle which can be obtained as:
\begin{equation}
    \theta^* = \arccos \left(\frac{\mathtt{tr}(\mathsf{R}^*_k)-1}{2}\right).
\end{equation}
$a^* = \phi^* / \theta^*$ is the rotation axis, and $a^{*\mathsf{H}}$ is its conjugate transpose. $[a^*]$ denotes the skew-symmetric matrix of $a^*$. $\widetilde{\mathsf{R}}^*$ is the rotation Jacobi defined as:
\begin{equation}\label{eq:rotation_jacobi}
    \widetilde{\mathsf{R}}^* = \frac{\sin \theta^*}{\theta ^*}\mathsf{Id} + \left( 1 - \frac{\sin \theta^*}{\theta^*}\right)a^* a^{*\mathsf{H}} + \left(\frac{1 - \cos \theta^*}{\theta^*}\right) [a^*].
\end{equation}
The gradient of the predicted surface geometry can be obtained as:
\begin{align}\label{eq:pvnpxi}
\frac{\partial V_{g,k}}{\partial \xi_{k,i}} = \frac{\partial V_{g,k}}{\partial \mathsf{T}_{g, k}} : \frac{\partial \mathsf{T}_{g, k}}{\partial \xi_{k,i}} = \frac{\mathtt{Im}\left(V_{g,k}^*\right)}{h},    \nonumber\\
\frac{\partial N_{g,k}}{\partial \xi_{k,i}} = \frac{\partial N_{g,k}}{\partial \mathsf{T}_{g, k}} : \frac{\partial \mathsf{T}_{g, k}}{\partial \xi_{k,i}} = \frac{\mathtt{Im}\left(N_{g,k}^*\right)}{h}.   
\end{align}
Here $\xi_{k,i}$ is the $i$-th perturbed component of $\xi_k$. CSFD tracks the initial perturbation $hi$ along the computation. As a result, the chain rule in Eq.~\eqref{eq:pvnpxi} is never explicitly evaluated (as opposed to AD), and we can conveniently calculate the gradient in an end-to-end manner. 

The derivative w.r.t. pixel coordinate $u$ is less intuitive as $u \in \mathbb{Z}^2$ is discrete. To this end, we generalize the domain of integer-value pixel index to $\mathbb{R}^2$ by bilinear interpolation.
If the vertex depth significantly differs from its neighbors: 
\begin{equation}
    \sum_{u' \in \mathcal{N}(u)} |D_k(u) - D_k(u')| \geq \delta,
\end{equation}
we do not perturb this coordinate to avoid noisy and misinformed derivatives.

\paragraph{Differentiable ICP}
As soon as the vertex and normal maps are ready, we estimate the camera pose by applying multi-level ICP between $\{V_k, N_k\}$ and $\{V_{g,k-1}, N_{g,k-1}\}$. The ICP objective function is the summation of all the point-to-plane distances for all the credible surface measures:
\begin{equation}\label{eq:icp}
E(\xi_k) = \sum \left\| \big(\mathsf{T}_{g, k} V_k(u) - V_{g,k-1}(\hat{u})\big)^\top \cdot N_{g,k-1}(\hat{u})\right\|.
\end{equation} 
Here $\hat{u}=\left(\mathsf{K}\mathsf{T}^{-1}_{g,k-1}(\widetilde{\mathsf{T}}_{g,k})_jV_k(u)\right)$. The function $\pi(\cdot)$ performs perspective projection. $(\widetilde{\mathsf{T}}_{g,k})_j$ is the camera pose at the $j$-th iteration during ICP and $(\widetilde{\mathsf{T}}_{g,k})_{j=0}$ is initialized with the previous frame pose. In original KF, the optimization of ICP requires linearization of the transformation matrix. Let $\alpha$, $\beta$, $\gamma$ be the three Euler angles, and assume that the pose change between adjacent frames is small. The transformation matrix is approximated by setting $cos(\theta) \approx 1$ and $sin(\theta) \approx \theta$ as:
\begin{equation}\label{eq:icp approx}
    \mathsf{T}_{g,k}=
    \begin{bmatrix}
        \mathsf{R}_k & t_k\\
        0 & 1
    \end{bmatrix} \approx
    \begin{bmatrix}
        1 & \alpha & -\gamma & t_x\\
        -\alpha & 1 & \beta & t_y \\
        \gamma & -\beta & 1 &t_z\\
        0 & 0 & 0 & 1
    \end{bmatrix}.
\end{equation}
This poses Eq.~\eqref{eq:icp} as a linear least squares problem in the form of $\mathsf{A}x=b$, which can be solved by various methods such as the Normal Equation and Singular Value Decomposition.

The ICP process in the dense SLAM system is slightly different from generic ICP. In KF, the ICP process is executed on vertex maps and normal maps, and the corresponding vertex $V_{g,k-1}(\hat{u})$ is computed by perspective projection instead of searching for the nearest point. Therefore, ICP becomes differentiable as long as we extend the discrete pixel coordinate to $\mathbb{R}^2$, as in the case of ray casting.
In general, differentiating an iterative optimization procedure is difficult. The computational graph is often unknown at the time of compiling as we do not know how many iterations will suffice. Most optimization techniques are sequential, and the results from previous iterations are needed in order to proceed. Therefore, it is a common practice in existing differentiable frameworks to assume a fixed iteration count for iterative procedures~\cite{hu2019chainqueen}. CSFD does not depend on the computational graph, and the derivative evaluation is agnostic on the global computation. Therefore, we can use any optimization algorithms to solve Eq.~\eqref{eq:icp} without worrying about the iteration count. To this end, we implemented a second-order ICP procedure using Newton's method at each ICP step. The Hessian can be conveniently computed via Eq.~\eqref{eq:csfd_hessian}, and we do not need to approximate trigonometric functions as in Eq.~\eqref{eq:icp approx}.

\paragraph{Surface Update}
KF updates the global fusion $\mathcal{M}$ according to the camera pose $\mathsf{T}_{g, k}$ and the depth map $D_k$. $\mathcal{M}$ represents a 3D TSDF volume. Let $p\in \mathbb{R} ^3$ be a voxel in $\mathcal{M}$ to be reconstructed. Its truncated signed distance $F_k(p)$ and weight $W_k(p)$ are updated via:
\begin{align}
    &F_k(p) =\frac{W_{k-1}(p)F_{k-1}(p) + W_{D_k}(p)F_{D_k}(p)}{W_{k-1}(p) + W_{D_k}(p)}, \nonumber\\
    &W_k(p) = W_{k-1}(p) + W_{D_k}(p).
\end{align}
Here, $F_{D_k}(p)$ and $W_{D_k}(p)$ are:
\begin{align}
    &F_{D_k}(p) = \Psi \left( \|\mathsf{K}^{-1}\dot{x}\|^{-1} \|t_{g,k} - p\| -L(x)\right), \label{eq:bi_interpolate}\\
    &W_{D_k}(p)=1,
\end{align}
and we have $x$ and $\Psi$ defined as:
\begin{equation}
x = \pi(\mathsf{K}\mathsf{T}_{g,k}^{-1} p), \quad
\Psi(\eta ) =
\begin{cases}
         \displaystyle \min\left(1, \frac{\eta}{\mu}\right) \mathtt{sgn}(\eta)\quad \text{if}\; \eta  \geqslant -\mu \\
         \mathtt{null} \quad \text{else}.
\end{cases}
\end{equation}
We use the bilinear interpolation function $L(x)$ in Eq.~\eqref{eq:bi_interpolate} to estimate the depth value instead of using the nearest-neighbor lookup in original KF to restore the smoothness of the function.

As in Eq.~\eqref{eq:lie}, we use $\xi_k $ to encode the pose matrix  $\mathsf{T}_{g,k}$ and promote it to the complex form $\xi_k^*$, which transforms $(F_k, W_k)$ into the complex quantities $(F_k^*, W_k^*)$. The gradient of the function $(F_k, W_k)\leftarrow \mathtt{surface}_{-}\mathtt{update}(\mathcal{M}_{k-1}, \mathsf{T}_{g,k},D_k)$ can be computed as:
\begin{equation}
    \frac{\partial F_k}{\partial \xi_{k,i}} = \frac{\mathtt{Im}\big(F_k^*\big)}{h},\quad\text{and}\quad
    \frac{\partial W_k}{\partial \xi_{k,i}} = \frac{\mathtt{Im}\big(W_k^*\big)}{h}.   
\end{equation}

\subsection{X-ElasticFusion and X-PointFusion}
ElasticFusion (EF) and PointFusion (PF) are two classic dense SLAM algorithms. Unlike KF, they use \emph{surfels} as the representation of the global map. The front-end odometry of EF is similar to that of PF. However, EF includes a back-end optimization part, introducing \emph{random ferns} for loop detection and map optimization through the deformation graph. This section focuses on the CSFD-based differentiation of the front-end odometry to jointly discuss those two methods. Other steps such as the camera pose estimation are dealt with in a similar way as in KF.

In EF/PF, $\mathcal{M}$ is an unstructured set of global model points. At time stamp $k$, each point $p_{g,k}$ is associated with the position ${v}_{g,k}$, normal ${n}_{g,k}$, radius $r_{g,k}$, confidence counter $c_{g,k}$ and time stamp $t_{g,k}$.
Given a new measurement $\{ I_k, D_k \}$, vertex map $V_k$ and normal map $N_k$ are computed similar as in KF, i.e., Eqs~\eqref{eq:vertex} and \eqref{eq:normal}. After that, the pixel is either, as a new point, inserted into $\mathcal{M}$ or merged with existing global model points. This includes two steps.

\begin{figure}
    \centering
    \includegraphics[width=\linewidth]{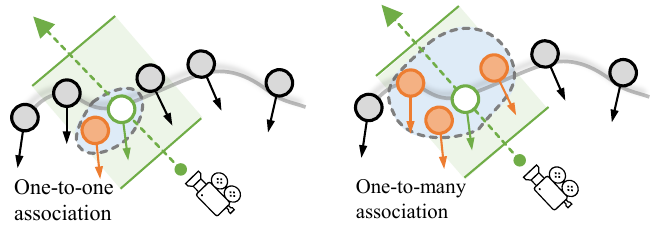}
    \Description[Figure 2: The illustration of our data association strategy.]{Left: one-to-one association in original ElasticFusion, Right: one-to-multiple association in our X-ElasticFusion. For each local point in current vertex map, we update all the qualified model points.}
    \caption{\textbf{An illustration of the one-to-one association (left) and our one-to-multiple association (right).} The associated global model points (gray dots) of camera ray (green dashed line) is colored in orange.}
    \label{fig:one to multi association}
  \end{figure}
\paragraph{Data association}
Given the estimated camera pose $\mathsf{T}_{g,k}$ and its intrinsic matrix $\mathsf{K}$, all global model points are rendered into an index map at the resolution $4$ by $4$ times higher than $D_k$, where the respective index $i$ is stored. Each pixel in $D_k$ thus corresponds up to 16 model points in this index map.  
We remove unqualified model point candidates for pixel $u$ if: 1) its depth distance along the viewing ray passing through $V_{g,k}(u)$ is beyond $\pm\delta_{depth}$; 2) its normal deviates from $N_{g,k}(u)$ by more than $\delta_{norm}$; 3) the distance to the latest measure $V_{g,k}(u)$ is farther than $\delta_{distance}$.
Because the number of points in $V_{g,k}(u)$ is smaller than the total number of global model points,
one-to-one point association as adopted in the original EF pipeline will generate a lot of vanished gradients at global model points. To reduce the sparsity of the gradient/Hessian, we use a one-to-many association strategy, which updates all the points with weight $w_{g, k}$ based on the Gaussian kernel function according to the distance from $V_{g,k}(u)$, i.e., see Fig.~\ref{fig:one to multi association}.

\paragraph{Surface update}
If pixel $u$ does not associate with existing model points, a new point is added to the global model. Otherwise, for all global model points corresponding to pixel $u$, the updated model point $p_{g,k}$ is computed as:
\begin{align}
    & v_{g, k} \leftarrow \frac{c_{g, k} v_{g, k} + w_{g, k} C_k(u) V_{g,k}(u)}{c_{g, k}+w_{g, k} C_k(u)},  \nonumber\\
    &n_{g, k} \leftarrow \frac{c_{g, k} n_{g, k} + w_{g, k} C_k(u) N_{g,k}(u)}{c_{g, k}+w_{g, k} C_k(u)}, \nonumber\\
    &c_{g, k} \leftarrow  c_{g, k} + w_{g, k} C_k(u),\quad  t_{g, k} \leftarrow k.
\end{align}

The confidence $C_k(u)$ is computed as in the vanilla PF:
\begin{equation}
    C_k(u) = e^{-\frac{\gamma^2}{0.72}}.
\end{equation}
Here $\gamma$ is the normalized radial distance of $V_k(u)$ from the camera center. Clearly, $p_{g,k}$ is a function of the depth map and camera pose. We promote $\xi_k$ to the complex form $\xi_k^*$, and $p_{g,k}$ is transformed into the complex form $p_{g,k}^*$ and the gradient can be computed with CSFD as:
\begin{equation}
    \frac{\partial p_{g,k}}{\partial \xi_{k,i}} = \frac{\mathtt{Im}\big(p_{g,k}^*\big)}{h}.
\end{equation}

\section{Evaluation}\label{sec:result}
We implemented X-KF on a desktop computer with an \texttt{intel} \texttt{i9} \texttt{13900KF} CPU and an \texttt{nvidia} \texttt{RTX} \texttt{4090} GPU. We implemented a (high-order) complex numerics library dedicated to CSFD differentiation, i.e., as discussed in \S~\ref{sec:csfd}. Some linear algebra computations are based on the \texttt{Eigen} library on the CPU using the \textsf{C++} version of our library as a template class, and most large-scale computations are implemented with \texttt{CUDA} on the GPU.  

We also built an active scanning system on a customized robot platform with a 6-DOF articulated arm holding an \texttt{intel} \texttt{RealSense} \texttt{D435i} RGBD sensor. The robot has a built-in computer running a ROS system, which provides a package to enable robot manipulation, such as navigation and arm actions. A laptop computer with an \texttt{intel i7 10750H} CPU and \texttt{nvidia 2070 GPU} to run X-EF and deep networks controls the robot through a wireless network.

\subsection{X-KF Evaluation}
\paragraph{Datasets} We conducted experiments on two datasets: \emph{ICL-NUIM dataset}~\cite{icl_nuim} and \emph{TUM RGB-D dataset}~\cite{tum_rgbd}. \emph{ICL-NUIM dataset} is a synthetic dataset with rendered RGBD images and ground truth camera poses. We selected two sequences including synthetic noise (\emph{lr kt1\_n, lr kt2\_n}) commonly used in the previous work to evaluate the performance of our pipeline. \emph{TUM RGB-D dataset} is a dataset captured by a \emph{Microsoft Kinect v1} with motion-captured camera poses as ground truth, which is widely used to evaluate the tracking accuracy of SLAM systems. We selected two sequences (\emph{fr1/desk, fr1/xyz}) for evaluation.

\paragraph{Comparison with AD method}
First, we demonstrate the advantages of CSFD compared to AD. We conduct tests on X-KF and $\nabla$KF as in $\nabla$SLAM~\cite{jatavallabhula2020GradSLAM}, which uses the backpropagation shipped with \texttt{PyTorch} as the major modality for computing differentiation. These two SLAM systems share the same parameters and the gradient $\frac{\partial F_k}{\partial \xi_{k,0}}$ is computed at each frame. We show the time performance and tracking accuracy in Table~\ref{table:comparison between AD and CSFD}. Due to the huge GPU memory consumption of $\nabla$KF, we can only run the first 100 frames of each sequence. On the other hand, CSFD is like a forward AD which does not need backpropagation. CSFD shows clear advantages in terms of both efficiency and accuracy.

\paragraph{Comparison with traditional SLAM}
To provide a more comprehensive and quantitative comparison, we compare the tracking accuracy between X-KF (at resolution of $512^3$) and two traditional SLAM systems, BundleFusion~\cite{dai2017bundlefusion} and ElasticFusion~\cite{whelan2015ElasticFusion} in Table~\ref{table:comparison between traditional methods}. X-SLAM does not achieve the state-of-the-art performance because those methods incorporate a back-end global optimization module, which is currently not integrated into X-SLAM. We also compare the time and memory performance of X-KF with original KF in Table~\ref{table:comparison between original KF}. The speed of X-KF is slightly slower than  original KF due to the additional computational costs for complex numbers. The spatial complexity of CSFD reformulation is $\mathbf{O}(N)$. In the worst case, the memory consumption is doubled due to the added imaginary part. For example, the main GPU memory consumption in KF is the TSDF volume which stores color (4 bytes), weight (4 bytes), and TSDF value (4 bytes) in each voxel. To use CSFD, we additionally use a float number (4 bytes) as the imaginary part of TSDF value, which results in approximately 33\% memory overhead. 
 However, we believe that this result is comparable and X-KF provides additional differential information.
{\small \fontfamily{ppl}\selectfont
\begin{table}
    \caption{\textbf{Time performance and tracking accuracy of X-KF and $\nabla$KF.} The tracking accuracy is measured with absolute trajectory error (ATE RMSE). Because the computational graph of AD method continuously grows in SLAM process, this experiment is performed on the first 100 frames for each sequence. The timing information (in millisecond) of the forward computation is also given in the parentheses. Even so, X-KF achieves around $10\times$ speed of $\nabla$KF and achieve similar tracking accuracy.}
    \renewcommand\arraystretch{1.3}
    \resizebox{\columnwidth}{!}{
    \begin{tabular}{c | c | c | c | c | c }
    \toprule[1pt]
        Sequence               & Res.    & \makecell{X-KF\\Time (ms)}   & \makecell{$\nabla$KF\\Time (ms)} & \makecell{X-KF \\ATE (m)} & \makecell{$\nabla$KF\\ATE (m)} \\ \hline
    \multirow{3}{*}{fr1\_desk} & $128^3$ &  8.4 (8.4)   & 118.9 (54.1)         & 0.026    & 0.025      \\
                               & $256^3$ &  9.5 (9.5)  & 154.6 (72.9)         &  0.022   & 0.024      \\
                               & $512^3$ &  11.2 (11.2) & 475.2 (137.9)        &  0.024   & 0.024      \\ \hline
    \multirow{3}{*}{fr1\_xyz}  & $128^3$ &  7.9  (7.9) & 113.3 (63.5)         & 0.018    & 0.020      \\
                               & $256^3$ &  10.1 (10.1) & 155.6 (81.2)         & 0.018    & 0.019      \\  
                               & $512^3$ &  11.5 (11.5) & 500.4 (160.3)        & 0.019    & 0.019      \\
    \bottomrule[1pt]
    \end{tabular}
    }
    \label{table:comparison between AD and CSFD}
\end{table} 
}
{\small \fontfamily{ppl}\selectfont
\begin{table}
    \caption{\textbf{Comparison of tracking accuracy.} Our tracking accuracy (ATE RMSE) is lower because we have not yet incorporated global optimization, leading to the drift along longer trajectories. The measurement is in meters.}
    \renewcommand\arraystretch{1.3}
    \resizebox{\columnwidth}{!}{
    \begin{tabular}{c | c | c | c }
    \toprule[1pt]
        Sequence               & \makecell{X-KF}   & \makecell{BundleFusion\\ \cite{dai2017bundlefusion}} & \makecell{ElasticFusion\\ \cite{whelan2015ElasticFusion}} \\ \hline
        {fr1\_desk} &  0.053  & 0.020        &  0.016         \\ 
        {fr1\_xyz}  &  0.020  & 0.010        &  0.011          \\
    \bottomrule[1pt]
    \end{tabular}
    }
    \label{table:comparison between traditional methods}
\end{table} 
}

{
{\small \fontfamily{ppl}\selectfont
\begin{table}[t]
    \caption{\textbf{Comparison of time/memory performance.} Both the time and memory performance are slightly worse than that of original KF due to the extra complex number computing process.}
    \renewcommand\arraystretch{1.3}
    \resizebox{\columnwidth}{!}{
    \begin{tabular}{c | c | c | c | c}
    \toprule[1pt]
        Sequence               & \makecell{X-KF \\ FPS}   & \makecell{KF \\ FPS} & \makecell{X-KF \\ Memory (MB)} & \makecell{KF \\ Memory (MB)} \\ \hline
        {fr1\_desk} &  82.9  & 101.5        &  2824   & 1937      \\ 
        {fr1\_xyz}  &  92.6  & 113.5        &  2234   & 1582       \\
    \bottomrule[1pt]
    \end{tabular}
    }
    \label{table:comparison between original KF}
\end{table} 
}
}

\paragraph{Simplified CSFD computation}
We have briefly introduced our complex library based on  simplified complex functions for CSFD in Section \ref{sec:csfd} and now we compare the time performance with other publicly available libraries. First, we compare the C++ version of our complex library with the C++ standard library on the CPU. We set $z_1=0.5+hi$, $z_2=-1.5+hi$ and evaluate some common functions. Each function is executed one million times, and the total time is presented in Table \ref{table:comparison of simplified CSFD on CPU}. Also, we compare the CUDA version of our complex library with the nvidia's library, libcu++ on GPU across the \emph{fr1/xyz} sequence of \emph{ TUM RGB-D dataset} for a $512^3$ voxel reconstruction. X-KF implemented by our complex library can achieve 92.6 FPS, while X-KF implemented by libcu++ only achieves 59.4 FPS, resulting in 1.55 times slower.

{\small \fontfamily{ppl}\selectfont
\begin{table}[t]
    \caption{\textbf{Computation time (unit: $ms$) of our complex numerics library and C++ STD library.} We test a variety of common mathematical operations, and the results show that our complex number numerics library is faster. And as the complexity of the operations increases, our improvement becomes more apparent.}
    \renewcommand\arraystretch{1.3}
    \begin{tabular}{c | c | c | c | c | c}
    \toprule[1pt]
    Methods& $z_1 \cdot z_2$ &  $z1 \mathbin{/} z2$ & $exp(z_1+z_2)$ & $sin(z_1+z_2)$ & $(z_1+z_2)^3$ \\ \hline
    STD    &  26.92 & 39.94 & 58.61 & 81.58 & 155.08 \\
    Ours   &  24.32 & 30.45 & 47.48 & 67.06 & 102.44 \\
    \bottomrule[1pt]
    \end{tabular}
    \label{table:comparison of simplified CSFD on CPU}
\end{table}
}

\begin{figure}[H]
    \centering
    \includegraphics[width=\linewidth]{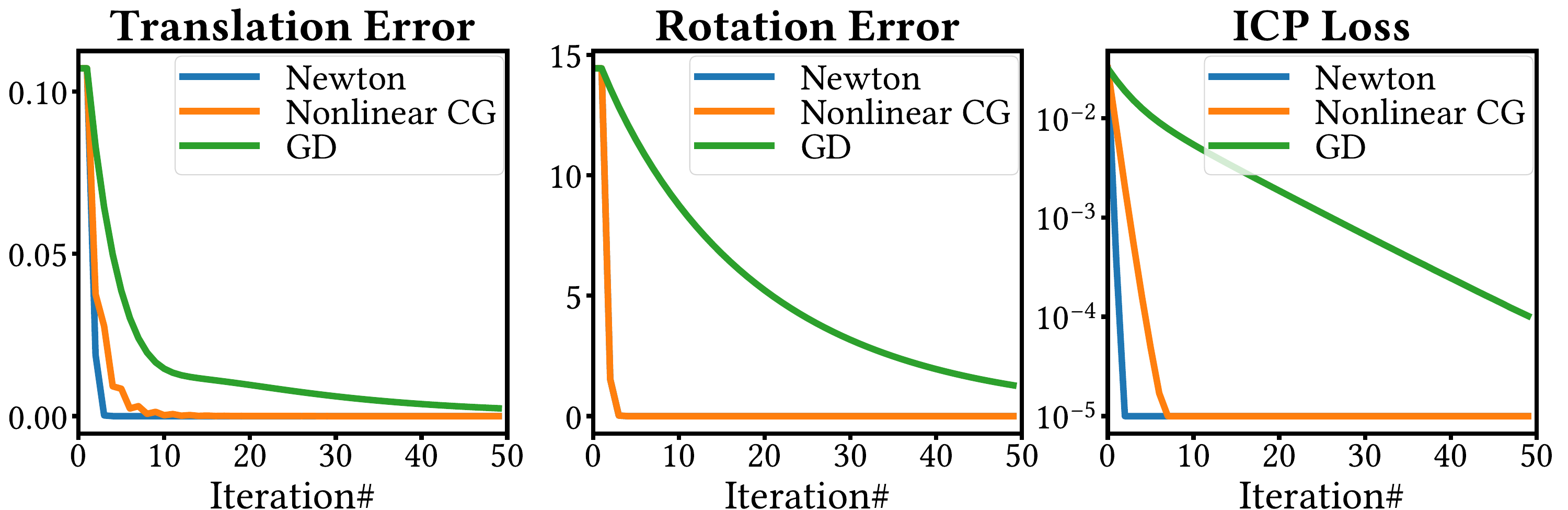}
    \Description[Figure 3: The optimization process of ICP based on Gradient Descent, Nonlinear-CG, and Newton’s method.] {Left: Translation Error, Middle: Rotation Error, Right: ICP loss. The results show that Newton's method have a faster convergence speed.}
    \caption{\textbf{The optimization process of ICP based on different optimization methods.} Translation error (unit: $m$), rotation error (unit: $^\circ$) and point-to-plane ICP loss are reported.}
    \label{fig:ICP optmization compare}
\end{figure}

\paragraph{High order ICP}
Finally, we test the performance of ICP based on different optimization algorithms for pose estimation. We project a depth image from \emph{lr kt1\_n} of \emph{ICL-NUIM dataset} to obtain the source point cloud, and the target point cloud is generated using a random transformation. We run point-to-plane ICP using three methods: Gradient Descent, Nonlinear-CG, and Newton's method. The optimization process is shown in Figure \ref{fig:ICP optmization compare} and the results show that higher-order optimization methods have a significantly faster convergence speed.

Also, we compare the tracking robustness of our X-KF and original KF on \emph{lr kt2\_n} of the \emph{ICL-NUIM dataset} with different frame step. It can be found that although the tracking robustness is similar when frame step is 1, as the frame step increases, the performance of X-KF surpasses that of  original KF. This is attributed to the following reasons. First, the convergence speed of Newton's method is higher than that of first-order optimization methods. Additionally, as the frame step increases, the pose differences between adjacent frames become larger, and the linear approximation of ICP in original KF works only when the pose differences between adjacent frames are small.  

\begin{figure}[t]
    \centering
    \includegraphics[width=\linewidth]{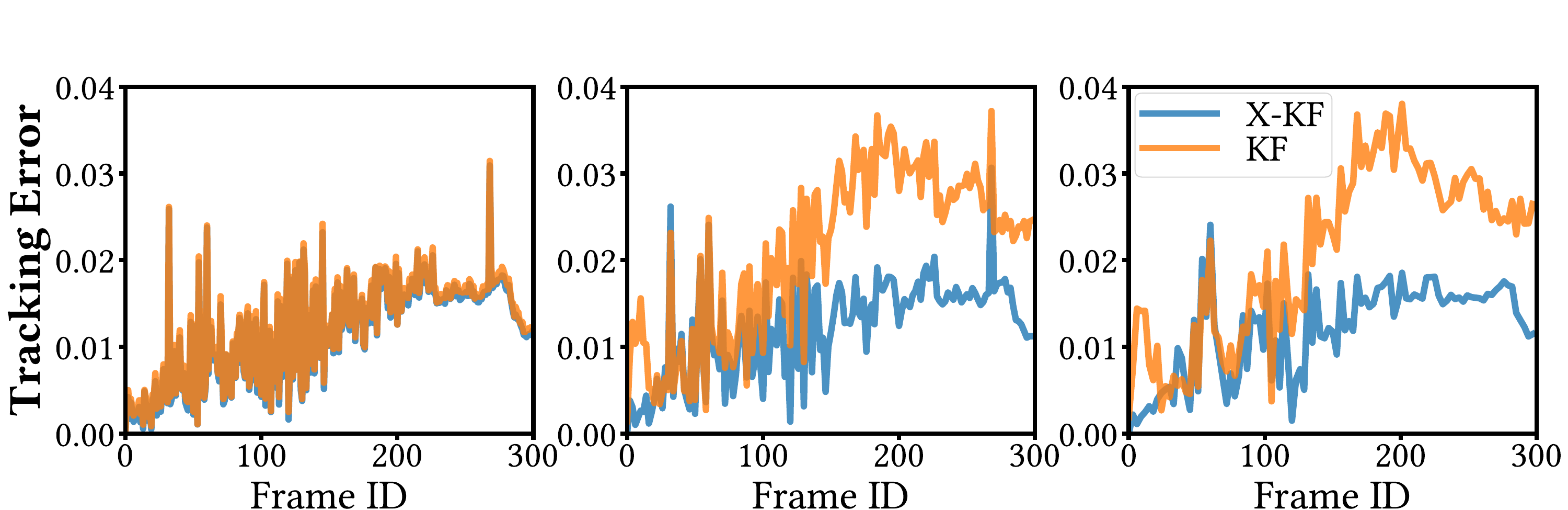}
    \Description[Figure 4: Comparison of the tracking accuracy between X-KinectFusion and original KinectFusion]{Left: frame step = 1, Mid: frame step = 2, Right: frame step = 3. The accuracy of high-order ICP surpasses that of original when the frame step increases.}
    \caption{\textbf{Absolute distance error (unit: $m$) during tracking.} Left: frame step = 1, Mid: frame step = 2, Right: frame step = 3. High-order ICP can achieve better tracking accuracy when the camera moves faster.}
    \label{fig:ICP tracking compare}
\end{figure}

\subsection{Camera Relocalization Based on X-KF} \label{sec:camera_relocalization}
A precise relocalization of the camera is a key component in vision tasks. Given multiple reference images with the corresponding camera poses $\{I_r, \mathsf{T}_{g,r}\}$, camera relocalization refers to estimating the camera pose $\mathsf{T}_{g,q}$ of any query image $I_q$.  {CSFD is very suitable for camera pose optimization because it only requires perturbations on a small number (i.e., 6) of parameters.} We show that an accurate camera relocalization can be conveniently obtained using depth maps based on X-KF. Let $\mathcal{M}_r$ be the reference TSDF volume. We compute a TSDF volume $\mathcal{M}_q$ such that $(F_q, W_q) \leftarrow \mathtt{surface}_{-}\mathtt{update}(\mathcal{M}_r, \mathsf{T}_{g,q}, D_q)$ for a query RGBD image $(I_q, D_q)$ and an estimated camera pose $\mathsf{T}_{g,q}$. Then an objective function can be defined as:
\begin{equation}
   E(\mathsf{T}_{g,q}) = \sum_{p}\left(F_r(p)-F_q(p)\right)^2. 
\end{equation}
The initial estimation for optimization can be obtained using any existing method. With X-KF, we can easily obtain the Hessian matrix and the Jacobian matrix of the objective function $E(\mathsf{T}_{g,q})$, and use an optimizer, i.e. Newton's method, to minimize $E(\mathsf{T}_{g,q})$. Since the TSDF values of the same voxel may vary at different viewpoints, we only use the images around the initial estimation to reconstruct $\mathcal{M}_r$.

\paragraph{Comparison} 
We compare our method with multiple state-of-the-art visual relocalization approaches. HLoc~\cite{sarlin2019Coarse} and PixLoc~\cite{sarlin2021Back} are feature-based methods for estimating the poses of the query images, which only require RGB images as input. VS-Net~\cite{huang2021vs} proposes scene-specific 3D landmarks and generates landmark segmentation maps to help with camera relocalization. DSM~\cite{tang2021Learning} is a scene-agnostic camera relocalization method that regresses the scene coordinates of the query images and the camera poses are solved by PnP algorithms. However, DSM requires the scene coordinates corresponding to the reference images. DSAC*~\cite{brachmann2021visual} is a combination of scene coordinate regression and differentiable RANSAC for end-to-end training and designs multiple workflows for different input data structures. SC-wLS~\cite{wuSCwLSInterpretableFeedforward2022} proposes a network to directly regress the camera pose based on the pipeline of DSAC*.

\paragraph{Experiments on 7-Scenes dataset} 
We start our evaluation on \emph{7-Scenes dataset}~\cite{seven_scenes} including small indoor scenes, which is commonly used to compare camera relocalization approaches. We use the aforementioned visual relocalization approaches to estimate the camera pose of the query image as the initial solution. For the following optimization, we implement both Gradient Descent and Newton's method. As Newton's method is sensitive to the initial solution, we set a threshold $\eta$ when applying  Newton's method for optimization. If the loss exceeds $\eta$, we use the gradient direction $-g$; instead, we use the Newton's direction $-H^{-1}g$. We report the statistical results in Table \ref{table:7-Scenes relocalization result} and the initial/optimized HLoc trajectories   in \emph{7-Scenes dataset}. The results indicate that our method is applicable not only to HLoc, PixLoc and SC-wLS, which only use RGB images for relocalization, but also improves the accuracy of the results for DSM and DSAC* that already utilize depth information. This improvement is particularly significant in \emph{Pumpkin}, \emph{redkitchen}, and \emph{stairs} where traditional methods do not perform well. In addition, we must point out the important role of higher-order differentials in the optimization process. It is evident that the Newton's method outperforms Gradient Descent in terms of convergence accuracy. On the other hand, we list the results of Gradient Descent based on the gradients computed by FD, and the results are worse than those of CSFD due to the low accuracy. We also report the optimization curves in Figure \ref{fig:camera reloc optimize curve} and the results indicate that Newton's method outperforms Gradient Descent in terms of convergence speed as well.

\paragraph{Experiments on our dataset} 

We also build a large-scale RGBD dataset to validate our method in larger scale environments. Our dataset consists of two outdoor buildings and two office rooms. For each scene, we conduct two scans using a \emph{Microsoft Kinect v2}. One scan reconstructed by BundleFusion is the reference set and the other is the query set. Because the depth images are not registered to the color images, we only evaluate HLoc, PixLoc and DSAC* (RGB mode) and each method is re-trained on the reference set. It should be noted that due to the issues such as drift during the scanning of large-scale scenes, we do not have the ground truth camera poses for the query set. Given the reference pointcloud $P_{g,r}$ and the query pointcloud $P_q$ corresponding to the depth map $D_q$, i.e., Eq~\eqref{eq:vertex}, we compute $e_q$ to measure the accuracy of the camera pose $\mathsf{T}_{g,q}$:

\begin{equation}
    e_q(\mathsf{T}_{g,q})=\frac{1}{|P_q|}\sum_{p_q\in P_q} \mathop{\min}_{p_{g,r}\in P_{g,r}}||\mathsf{T}_{g,q} p_q - p_{g,r}||.
\end{equation}

We consider  the frames with $e_q > 0.1m$ as outliers and exclude them from  statistical analysis. We report the result in Table \ref{table:our dataset reloc result}. Similar to the results in \emph{7-Scenes dataset}, our optimization method can still improve the accuracy of camera relocalization even in complex and large-scale environments. We also show the relocalization trajectories initialized from PixLoc and the reference model in Figure \ref{fig:camera reloc traj on ourdataset}. 

{\small 
\begin{table}
    \caption{\textbf{Camera relocalization results (unit: $cm$) in our large-scale datasets.} We use the nearest neighbor distance between the projected query depth map and the reference model. The initial/GD optimized/Newton optimized results are reported.}
    \renewcommand\arraystretch{1.3}
    \begin{tabular}{c|c|c|c}
    \toprule[1pt]
    Scene     & HLoc                    & PixLoc                  & DSAC*   \\ \hline
    Outside A & 2.34/1.46/\textbf{1.33} & 2.75/1.75/\textbf{1.58} & 3.36/1.95/\textbf{1.78}\\ 
    Outside B & 1.90/1.53/\textbf{1.47} & 2.20/1.75/\textbf{1.67} & 3.38/2.33/\textbf{2.17} \\ 
    Office A  & 3.03/1.84/\textbf{1.72} & 3.19/2.12/\textbf{1.99} & 3.30/2.43/\textbf{2.20} \\ 
    Office B  & 3.02/2.32/\textbf{2.21} & 3.69/3.15/\textbf{3.02} & Not converge       \\ 
    \bottomrule[1pt]
    \end{tabular}
    \label{table:our dataset reloc result}
\end{table}
}

{\small 
    \begin{table*}[tb]
    \caption{\textbf{Camera relocalization results in \emph{7-Scenes dataset}.} We report the median translation (unit: $cm$), rotation (unit: $^\circ$) errors and the average recall at $(5cm, 5^\circ)$. Whether these methods consider depth information or not, our method can further enhance accuracy, especially Newton's method based on higher-order derivatives.}
    \renewcommand\arraystretch{1.4}
    \resizebox{\textwidth}{!}{
    \begin{tabular}{c | c | c | c| c| c| c| c| c}
    \toprule[1pt]
        Method &   Optimization     & Chess & Fire & Heads & {Office} & Pumpkin & Redkitchen & Stairs \\ \hline
    \multirow{4}{*}{\makecell{HLoc\\ \cite{sarlin2019Coarse}}}   & Init   & 2.41/0.85/92.3            & 2.28/0.91/89.3           & 1.08/0.74/96.3            & 3.12/0.90/76.7             & 4.83/1.27/51.8              & 4.21/1.39/58.9                 & 5.23/1.47/47.3             \\
                            &  {GD (FD)}     &  {2.32/0.82/93.3}            &  {2.22/0.89/90.9}           &  {1.02/0.75/97.0}            &  {3.09/0.91/77.8}             &  {4.52/1.26/54.4}              &  {4.02/1.31/61.2}                 &  {5.25/1.43/45.4}             \\
                            & GD     & 1.68/0.56/98.8            & 1.94/0.78/93.3           & 0.90/0.61/98.6            & 2.55/0.80/86.1             & 3.89/1.03/61.0              & 3.26/1.23/74.8                 & 4.96/1.27/50.7             \\
                            & Newton & \textbf{1.15/0.37/99.3}   & \textbf{1.19/0.48/99.0}  & \textbf{0.79/0.47/99.0}   & \textbf{1.46/0.47/99.0}    & \textbf{2.10/0.69/74.8}     & \textbf{2.37/0.86/83.3}        & \textbf{2.62/0.70/81.2}    \\ \hline
    \multirow{4}{*}{\makecell{PixLoc\\ \cite{sarlin2021Back}}} & Init   & 2.37/0.81/91.0            & 1.90/0.78/86.5           & 1.26/0.86/87.4            & 2.57/0.79/81.7             & 4.11/1.19/59.0              & 3.41/1.23/67.5                 & 4.49/1.22/53.4             \\
                            &  {GD (FD)}     &  {2.31/0.79/91.4}            &  {1.86/0.76/86.7}           &  {1.13/0.83/87.5}            &  {2.53/0.79/82.8}             &  {3.37/1.19/68.1}             &  {3.33/1.22/68.7}                 &  {4.69/1.19/53.2} \\
                            & GD     & 1.65/0.55/96.0            & 1.61/0.74/87.5           & 1.01/0.65/90.0            & 2.15/0.71/85.3             & 3.43/0.99/66.7              & 2.93/1.11/75.8                 & 4.39/1.06/55.5             \\   
                            & Newton & \textbf{1.11/0.37/96.5}   & \textbf{1.15/0.46/89.2}  & \textbf{0.82/0.50/96.4}   & \textbf{1.47/0.46/95.0}    & \textbf{2.10/0.69/73.7}     & \textbf{2.50/0.91/79.1}        & \textbf{3.18/0.75/65.4}    \\ \hline
    \multirow{4}{*}{\makecell{VS-Net\\  {\cite{huang2021vs}}}}  & Init   & 1.57/0.52/98.9            & 1.91/0.82/96.2           & 1.20/0.72/98.7            & 2.14/0.60/91.5             & 3.84/1.05/64.1              & 3.59/1.08/72.7                 & 2.79/0.77/93.4             \\
                            &  {GD (FD)}     &  {1.54/0.53/99.0}           &  {1.86/0.78/96.5}           &  {0.96/0.74/99.4}             &  {2.07/0.62/92.5}             &  {3.57/1.05/65.7}              &  {3.44/1.03/74.6}                 &  {2.78/0.77/92.9}             \\
                            & GD     & 1.21/0.41/\textbf{99.6}            & 1.46/0.67/98.5           & 0.77/0.61/\textbf{100}             & 1.78/0.59/95.1             & 3.44/1.01/67.3              & 2.82/1.02/85.5                 & 2.58/0.74/94.5             \\
                            & Newton & \textbf{0.96/0.35/}99.5   & \textbf{0.92/0.41/99.4}  & \textbf{0.48/0.40/100}    & \textbf{1.27/0.41/98.5}    & \textbf{2.62/0.75/76.5}     & \textbf{2.32/0.87/87.0}        & \textbf{1.93/0.60/98.2}    \\ \hline
    \multirow{4}{*}{\makecell{DSAC*\\ \cite{brachmann2018DSAC}}}  & Init   & 1.01/0.44/98.9            & 1.12/0.53/99.1           & 0.98/0.87/\textbf{100}             & 1.19/0.48/\textbf{99.9}             & 1.97/0.67/80.9              & 2.09/0.83/92.1                 & 2.62/0.78/91.2             \\
                            &  {GD (FD)}     &  {0.99/0.43/99.0}            &  {1.13/0.54/98.9}          &  {0.98/0.81/\textbf{100}}             &  {1.26/0.48/99.1}            &  {2.01/0.66/80.4}              &  {2.15/0.77/91.5 }                &  {2.67/0.78/91.6}             \\
                            & GD     & 0.85/\textbf{0.34}/99.4            & 0.95/0.45/99.3           & 0.76/0.63/\textbf{100}             & \textbf{1.14}/0.44/99.4             & 1.67/0.51/86.4              & 1.64/0.54/93.8                 & 2.58/0.78/\textbf{94.2}             \\
                            & Newton & 0.93\textbf{/0.34/99.5}   & \textbf{0.85/0.37/99.7}  & \textbf{0.50/0.42/100}    & 1.24/\textbf{0.41}/99.7    & \textbf{1.20/0.38/89.6}     & \textbf{1.04/0.36/98.4}        & \textbf{1.89/0.60/}93.6    \\ \hline
    \multirow{4}{*}{\makecell{DSM\\ \cite{tang2021Learning}}}    & Init   & \textbf{2.01}/0.70/94.7            & 2.62/0.82/0.88           & 1.47/0.84/95.4            & 3.53/0.82/70.7             & 5.21/1.11/47.5              & 4.80/1.12/52.2                 & 4.89/1.36/50.9             \\
                            &  {GD (FD)}    &  {2.07/0.76/94.3}            &  {2.60/0.87/88.2}           &  {1.43/0.92/95.5}            &  {3.46/0.79/72.3}             &  {4.87/1.20/51.0}              &  {4.71/1.15/53.6}                 &  {4.86/1.16/67.2}             \\
                            & GD     & 2.13/0.66/96.6            & 2.04/0.75/95.2           & 1.26/0.78/97.2            & 3.01/0.82/79.2             & 4.08/1.09/61.3              & 3.96/1.02/64.8                 & 3.27/0.86/76.0             \\
                            & Newton & 2.02\textbf{/0.60/96.8}   & \textbf{1.71/0.67/96.1}  & \textbf{1.09/0.71/97.5}   & \textbf{2.72/0.76/83.4}    & \textbf{3.06/0.86/74.6}     & \textbf{2.96/0.83/75.9}        & \textbf{2.66/0.62/82.5}    \\ \hline
    \multirow{4}{*}{\makecell{SC-wLS\\ \cite{wuSCwLSInterpretableFeedforward2022}}} & Init   & 3.03/0.76/78.9            & 4.19/1.08/55.0           & 2.86/1.92/60.7            & 5.18/0.86/0.48             & 7.29/1.27/28.7              & 8.27/1.44/26.4                 & 12.1/2.43/15.7             \\
                            &  {GD (FD)}    &  {2.75/0.82/80.2}            &  {3.91/1.15/56.3}           &  {2.80/1.84/60.1}            &  {4.52/0.96/54.7}             &  {6.72/1.41/36.4}              & {7.45/1.55/32.3}                 &  {11.9/2.43/19.0}             \\
                            & GD     & 1.45/0.50/89.7            & 2.28/0.90/65.4           & 1.41/1.12/\textbf{61.3}            & 2.66/0.79/76.9             & 4.65/1.24/52.7              & 5.01/1.44/49.7                 & 11.2/2.22/26.8             \\
                            & Newton & \textbf{1.04/0.38/90.9}   & \textbf{1.22/0.52/70.9}  & \textbf{0.95/0.75/61.3}   & \textbf{1.56/0.51/83.4}    & \textbf{3.27/0.85/59.8}     & \textbf{3.24/1.06/62.7}        & \textbf{9.55/1.72/39.1}    \\ 
    \bottomrule[1pt]
    \end{tabular}
    }
    \label{table:7-Scenes relocalization result}
\end{table*}
}

\begin{figure*}[htb]
    \centering
    \includegraphics[width=\textwidth]{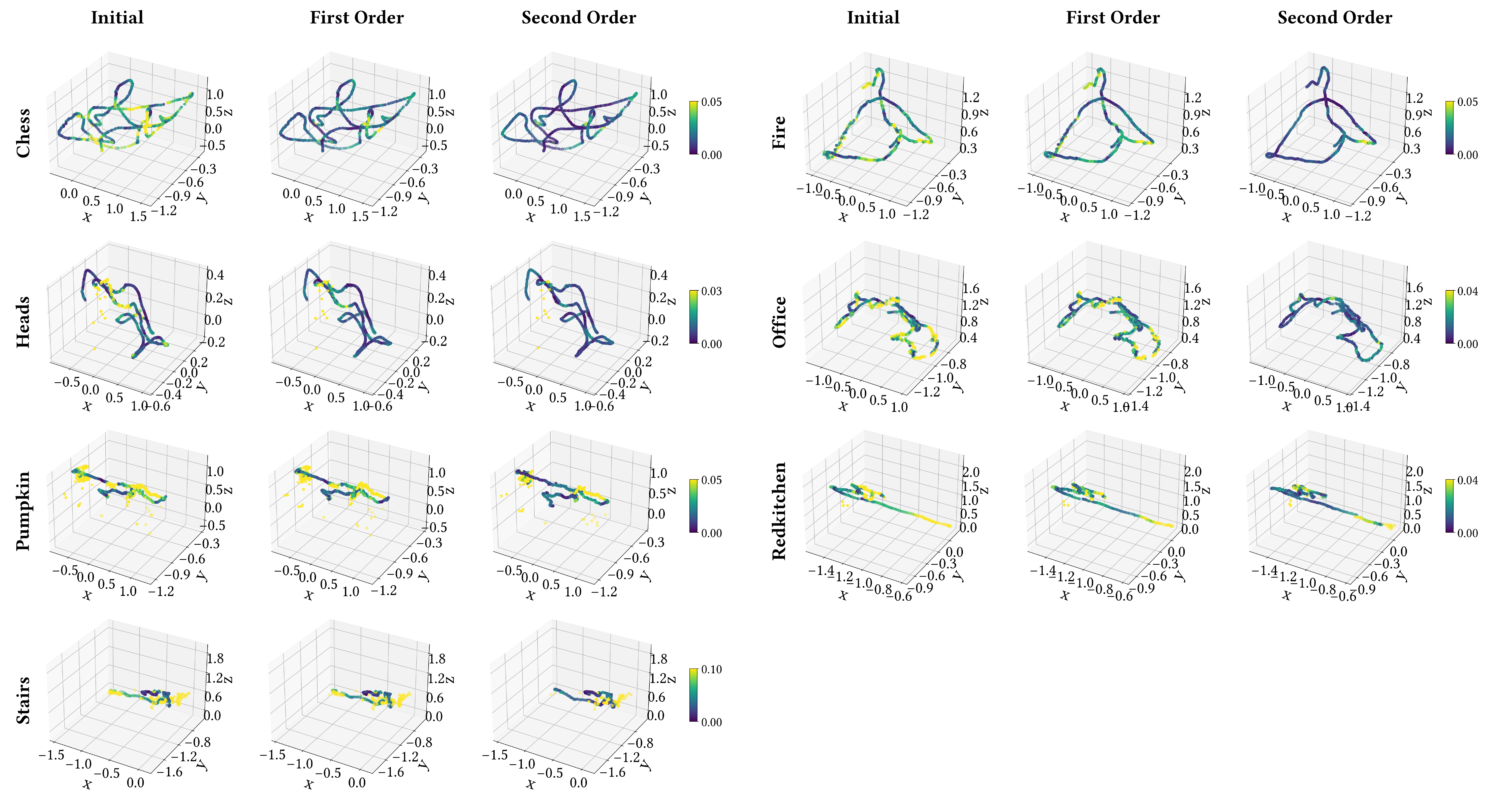}
    \Description[Figure 5: Camera relocalization trajectories in 7-Scenes dataset for HLoc.]{For each scene we show the trajectory of the initial solution of HLoc, the trajectory optimized by Gradient Descent and the trajectory optimized by Newton's method. All trajectories are colored by the absolute distance error. The results show that our optimization pipeline can increase the relocalization accuracy and higher-order optimization can achieve better results.}
    \caption{\textbf{Camera relocalization trajectory in \emph{7-Scenes dataset} for HLoc.} For all scenes, our optimization method based on X-KF can significantly enhance the relocalization accuracy. The results also demonstrate the importance of higher-order optimization, which is difficult for AD method.} 
    \label{fig:camera reloc traj on 7-Scenes}.
\end{figure*}

\begin{figure*}[htb]
    \centering
    \includegraphics[width=\textwidth]{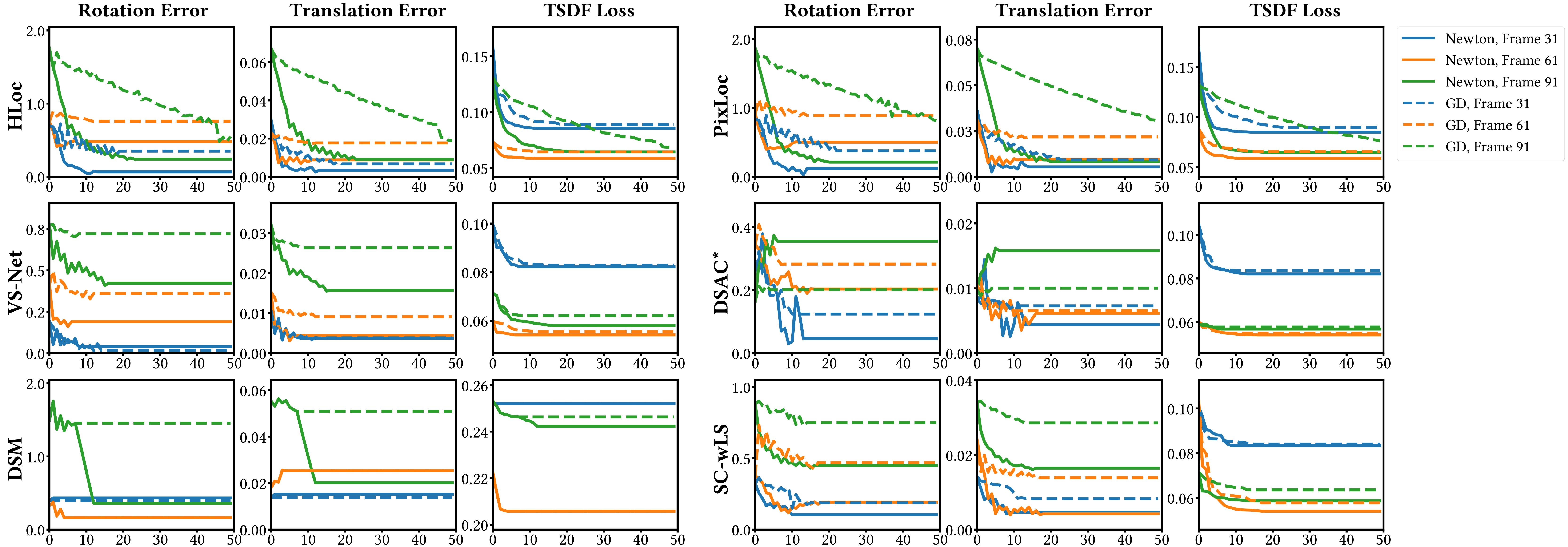}
    \Description[Figure 6: Optimization curves on the chess scene in 7-Scenes dataset.]{Three frames are optimized 50 iterations in Gradieent Denscent and Newton's method, and the optimization curves on rotation error, translation error and TSDF loss are shown. Newton's method achieves better results in both convergence speed and final accuracy}
    \caption{\textbf{Optimization curves on the chess scene in \emph{7-Scenes dataset}.} Three frames (No. 31, No. 61, No. 91) are optimized 50 iterations in Gradient Descent and Newton's method. Not only does Newton's method achieve a faster convergence speed, but it also prevents falling into local minima.
    }
    \label{fig:camera reloc optimize curve}
\end{figure*}

\begin{figure*}[htb]
    \centering
    \includegraphics[width=\textwidth]{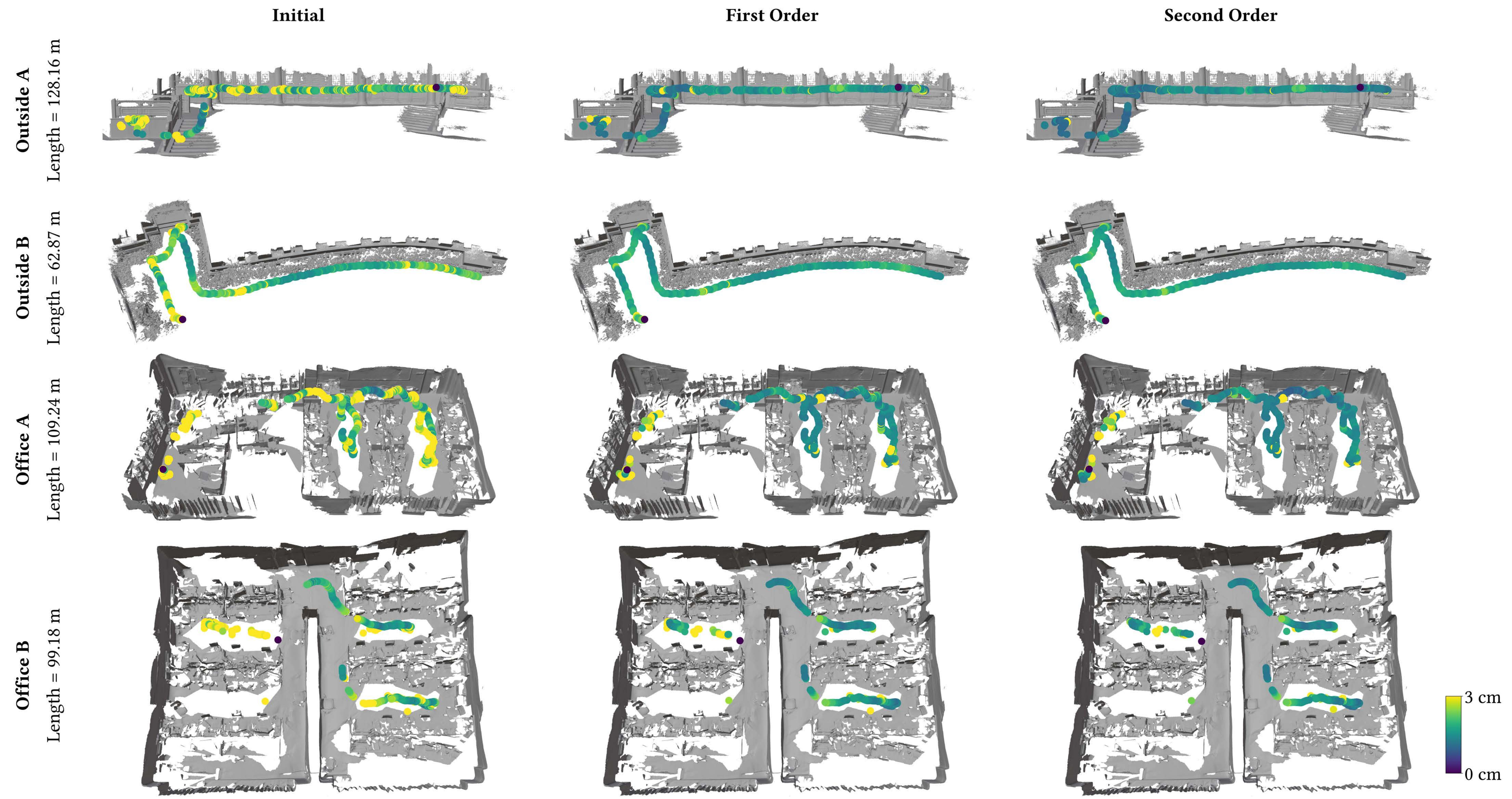}
    \Description[Figure 7: Camera relocalization trajectories in our large-scale dataset for PixLoc.]{We show the results of two outdoor buildings and two offices room. In each row, Left: the initial trajectory from PixLoc: Middle, the trajectory optimized by Gradient Descent, Right: the trajectory optimized by Newton's method. All trajectories are colored by the error defined in Equation 35.}
    \caption{\textbf{Camera relocalization trajectories in our large-scale dataset for PixLoc.} The trajectories are colored by relocalization error. Please note that we have discarded outliers with $e_q>0.1m$, which occur more frequently in indoor scenes because of local repetition.}
    \label{fig:camera reloc traj on ourdataset}
\end{figure*}

\subsection{Robot Active Scanning Based on X-EF}
Active scanning and online reconstruction by robots is another important downstream task of SLAM. We demonstrate that, by leveraging X-EF, we can combine this task with neural networks to achieve better performance. We follow the pipeline in \cite{liu2018object}:
\begin{enumerate}
    \item \emph{Object Segmentation.} Robots perform real-time scene reconstruction using X-EF and then segment the point cloud to obtain a series of objects.
    \item \emph{Target Object Decision.} Robots evaluate these objects based on a set of criteria such as distance, size, orientation, etc., and choose the optimal scanning target, called the Next Best Object (NBO)
    \item \emph{Target View Decision.} Once the NBO is determined, the  object point cloud $P_O$ from X-EF is sent to a pre-trained PointNet~\cite{qi2017pointnet}  network to predict a label score $S$, and the Next Best View (NBV) is defined as the viewport that maximizes the recognition score.
    \item \emph{Repeating Scans.} Once the NBV is determined, the robot is instructed to move to the new viewpoint. PointNet is then re-evaluated, assigning scores to the objects based on the new perspective. This process continues until the recognition score exceeds the threshold, and the system returns to select another NBO.
\end{enumerate}
In our system, the NBV selection process aims to maximize the recognition score $S$. An objective function can be defined as:
\begin{align}
   &E(\mathsf{T}_{g,k}) = -S = -\mathtt{point\_net}(P_O). 
\end{align}
Using X-EF, we can calculate the differentiation of $S$ w.r.t. the camera pose:
\begin{equation}
    \frac{\partial S}{\partial \xi_{k,i}} = \sum_{p_O\in P_O} \frac{\partial S}{\partial p_O}: \frac{\partial p_O}{\partial \xi_{k,i}}.  
\end{equation}
Here the network gradient $\frac{\partial S}{\partial p_O}$ is computed using  AD method, while the SLAM gradient  $\frac{\partial p_O}{\partial \xi_{k,i}}$ is calculated by X-EF. Similar to camera relocalization in Section \ref{sec:camera_relocalization}, CSFD can quickly calculate the SLAM gradient for the 6DOF camera pose. Since computing higher-order differentials is challenging with AD method, and the dimension of $\xi$ is only 6, we employ finite differences based on gradients to compute the Hessian matrix.

\begin{figure}[htb]
    \centering
    \includegraphics[width=\linewidth]{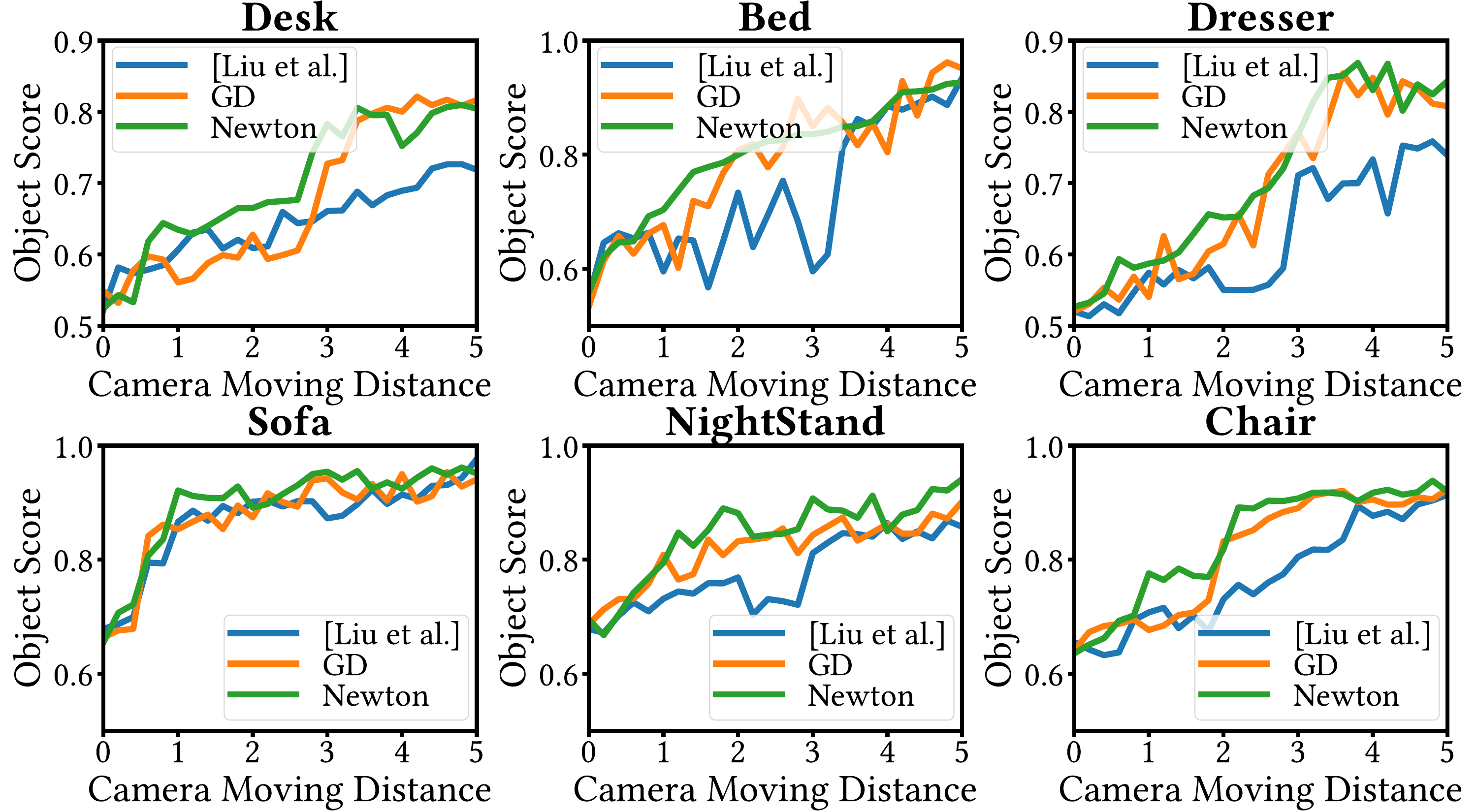}
    \Description[Figure 8: Recognition score during robot activate scanning.]{We evaluate the performance of NBV-based object recognition methods on 6 categories (Desk, Bed, Dresser, Sofa, NightStand and Chair). Our method outperforms database-based method in most categories.}
    \caption{\textbf{Recognition score during robot activate scanning.} We record the object recognition score given by PointNet every 0.2$m$. On most objects, our method can achieve better results than the database-based method.
    }
    \label{fig:nbv compare}
\end{figure}

\paragraph{Performance of object recognition}
We first evaluate the performance of  NBV-based single object recognition. To provide a quantitative evaluation, we compare our method with \cite{liu2018object} in the robotic simulation environment of \emph{Gazebo} running on top of ROS. We pre-train a PointNet network on \emph{ModelNet40 dataset}~\cite{modelnet40} and choose several object models from the test split to evaluate. For each model, we scale it to an appropriate size and add random perturbations to the horizontal orientation and position. We limit the scanning trajectory length to $5m$, and record the recognition score corresponding to the ground truth label every $0.2m$. 20 models are selected for each category, and the average scores are reported in Figure \ref{fig:nbv compare}. Please note that we compare the moving distance for achieving the same level of recognition accuracy, which is more appropriate because we find there may be a significant distance between two viewpoints. Our results are better than those of \cite{liu2018object} in most categories. On the other hand, although both first-order and second-order optimization end up with similar scores in most cases, second-order optimization is more stable and converges faster.

\begin{figure*}[htb]
    \centering
    \includegraphics[width=\textwidth]{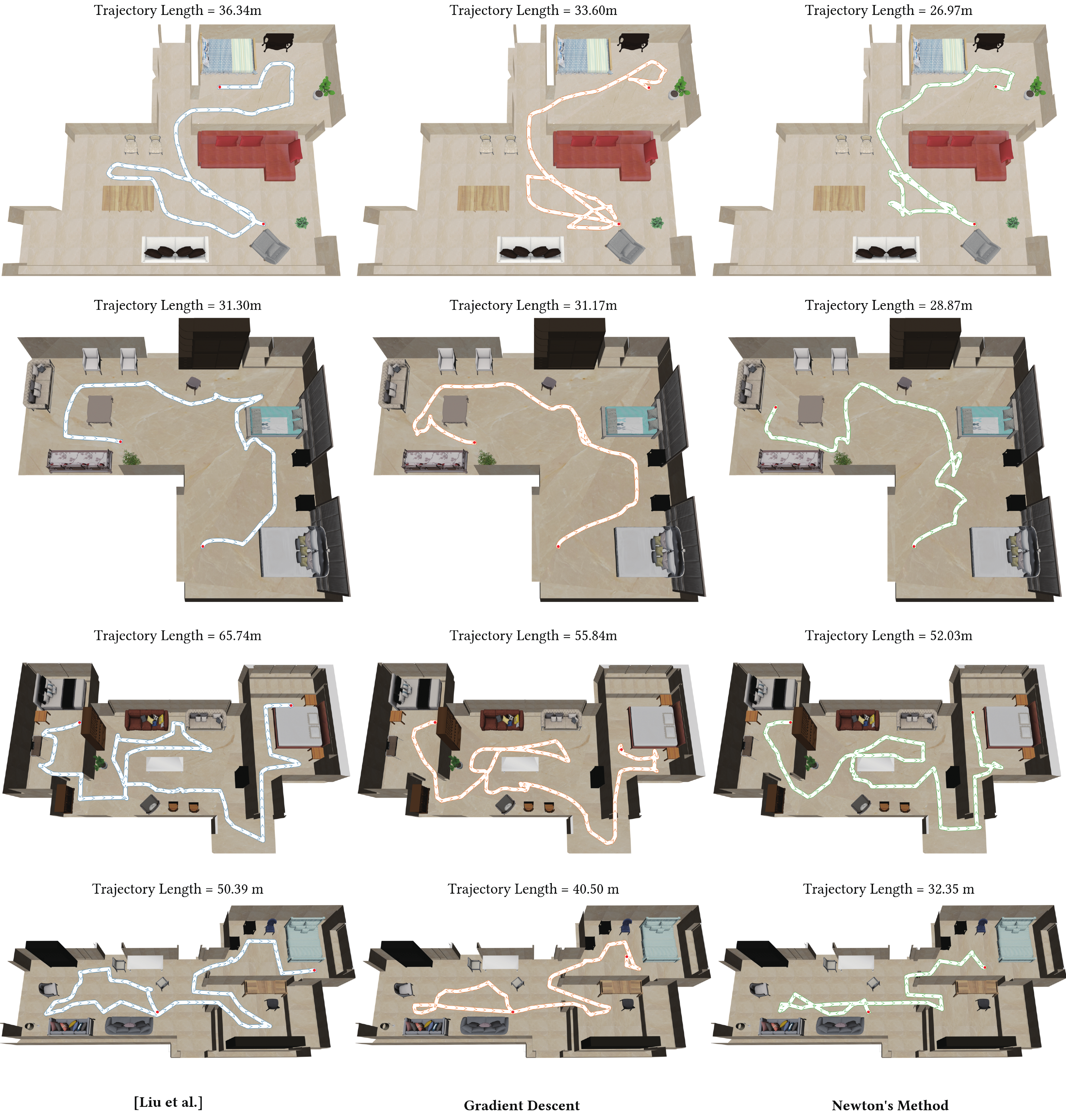}
    \Description[Figure 9: Robot active scanning in virtual scenes.]{We show the scanning trajectories on 4 virtual scenes. In each row, Left: trajectories of database-based method, Middle: trajectories of our method with Gradient Descent, Right: trajectories of our method with Newton's method. The robot starts from the same position and stops when all objects are recognized. The results show that our method achieves shorter paths and the high-order optimization can solve the problem of overshoot.}
    \caption{\textbf{Robot active scanning in virtual scenes.} The robot starts from the same position for active scanning, and the moving trajectory is drawn on the map. We also report the trajectory length.
    }
    \label{fig:virtual robot scan}
\end{figure*}

\paragraph{Performance of virtual scene scanning} 
Figure \ref{fig:virtual robot scan} shows robot activate scanning process in some virtual scenes from \emph{3D-FRONT dataset}~\cite{fu20213d}. It should be noted that since \cite{liu2018object} did not provide the 3D model database, we build our own model database using \emph{ModelNet40 dataset} for comparison. All methods are compared for achieving the same level of recognition completeness. The scanning process is stopped when all objects are recognized (with score > 0.8).  We draw the scanning trajectory, and the trajectory length is also reported. It can be observed that our method achieves simpler and shorter paths. Although Gradient Descent can get a good moving direction, it is limited by overshoot, which sometimes makes the robot move back and forth. In contrast, Newton's method can solve this problem, improving the scanning path.

\begin{figure*}[htb]
    \centering
    \includegraphics[width=\textwidth]{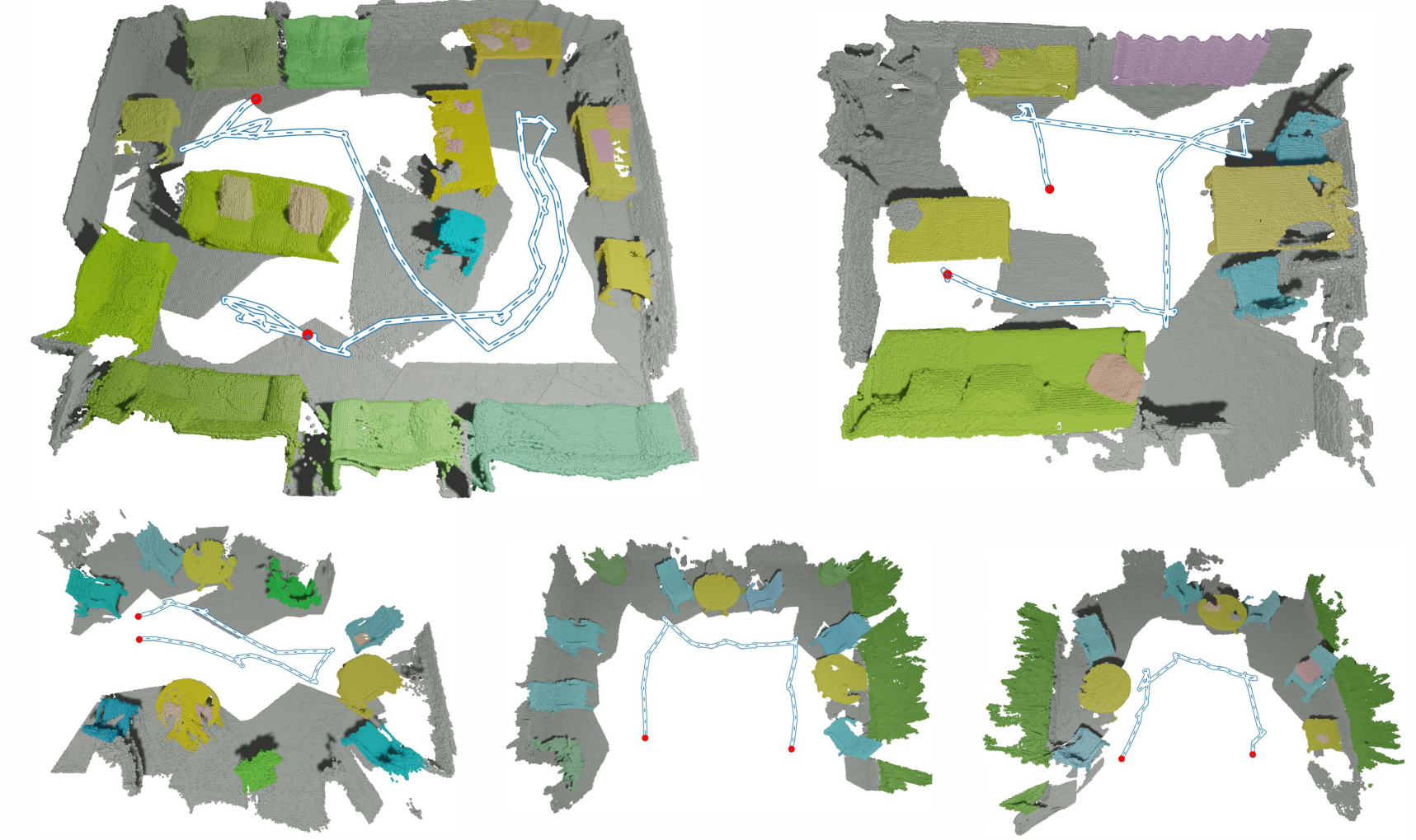}
    \Description[Figure 10: Segmentation and reconstruction results in real scene scanning.]{The results of a resting room and a living room are shown in the first row, and the results of three outdoor terraces are shown in the second row. Based on our method, the robot successfully recognizes and segments most objects in all scenes.}
    \caption{\textbf{Segmentation and reconstruction results in real scene scanning.}}
    \label{fig:real robot scan}
\end{figure*}

\paragraph{Performance of real scene scanning}
We test our system by scanning five unknown scenes, including a resting room, a living room and three outdoor terraces. Figure \ref{fig:real robot scan} shows the reconstruction results in these scenes. The results show that the robot can actively complete scene reconstruction and recognize most objects.

\section{Conclusion}
We present a real-time and differentiable dense SLAM system based on CSFD. By adding perturbations to variables of interest, the numerical differentiations can be computed in efficiently during the SLAM process. We propose X-KF and X-EF, which are the real-time differentiable versions of the two classic SLAM systems, and demonstrate that CSFD is faster and uses less memory compared to auto differentiation. Based on X-SLAM, we introduce task-aware optimization frameworks for two downstream tasks, camera relocalization and robot active scanning. Thanks to the ability of X-SLAM to compute high-order derivatives, our optimization framework shows superior performance in a variety of public datasets and difficult real-world environments. As CSFD requires perturbing variables one by one to calculate derivatives, our method mainly focuses on pose optimization over a smaller number of degrees of freedom. It remains challenging to carry out high-order differential calculations and high-dimension degrees of freedom in the SLAM system.
%
\begin{acks}
The authors would like to thank the reviewers for their insightful comments. This work is supported by NSF China (No. U23A20311 \& 62322209), the XPLORER PRIZE, the 100 Talents Program of Zhejiang University, and NSF under grant numbers of
2301040, 2008915, 2244651, 200856.
\end{acks}

\bibliographystyle{ACM-Reference-Format}
\bibliography{ref}

\end{document}